\documentclass[11pt,a4paper]{article}
\usepackage{times,latexsym}
\usepackage{url}
\usepackage[T1]{fontenc}
\usepackage{comment}

\usepackage[acceptedWithA]{tacl2021v1}

\usepackage{xspace,mfirstuc,tabulary}

\newif\iftaclinstructions
\taclinstructionsfalse 
\iftaclinstructions
\renewcommand{\confidential}{}
\renewcommand{\anonsubtext}{(No author info supplied here, for consistency with
TACL-submission anonymization requirements)}
\newcommand{\instr}
\fi

\iftaclpubformat 

\else

\fi

\usepackage{latexsym}
\usepackage{multirow}

\usepackage{microtype}
\usepackage{tikz} 

\usepackage[noabbrev,capitalize,nameinlink]{cleveref}
\crefformat{section}{\S#2#1#3}  
\crefname{section}{\S}{\S\S}
\Crefname{section}{\S}{\S\S}
\crefname{table}{Table}{}
\crefname{figure}{Figure}{}
\crefname{algorithm}{Algorithm}{}
\crefname{equation}{Equation}{}
\crefname{appendix}{Appendix}{}
\crefname{thm}{Theorem}{}
\crefname{prop}{Proposition}{}
\crefname{cor}{Corollary}{}
\crefname{observation}{Observation}{}
\crefname{assumption}{Assumption}{}
\crefformat{section}{\S#2#1#3}
\creflabelformat{equation}{#2\textup{#1}#3}

\usepackage{graphicx}
\usepackage{todonotes}
\usepackage{booktabs}
\usepackage{array}
\usepackage{longtable}
\usepackage{arydshln}
\usepackage{subcaption}
\setuptodonotes{fancyline, color=green!30, size=footnotesize}
\newcommand{\spheading}[2][7em]{
  \rotatebox{90}{\parbox{#1}{\centering #2}}}

\newcommand{\textttt}[1]{{\small \texttt{#1}}}
\newcommand{\opensourceurl}{{\scriptsize\url{https://github.com/facebookresearch/worldsense}}}

\newcommand{\WS}{WorldSense}

\title{\WS{}: A Synthetic Benchmark for Grounded Reasoning\\in Large Language Models}

\author{
  Youssef Benchekroun$^1$\Thanks{Shared first authorship} \and Megi Dervishi$^{1,3}$$^*$ \and Mark Ibrahim$^1$$^*$
  \AND Jean-Baptiste Gaya$^1$ \and Xavier Martinet$^1$ \and Grégoire Mialon$^1$ \and Thomas Scialom$^1$
  \AND Emmanuel Dupoux$^{1,2}$\Thanks{Shared senior authorship} \and Dieuwke Hupkes$^1$$^\dagger$ \and Pascal Vincent$^{1,4}$$^\dagger$
  \\
  \ \\
  $^1$Meta $^2$EHESS \\
  $^3$LAMSADE, Université Paris Dauphine-PSL,
  $^4$Mila, Université de Montréal-DIRO
  \\
  \texttt{\{youssefb,megidervishi,marksibrahim,dieuwkehupkes,dpx,pascal\}}\\
  \texttt{@meta.com}
}

\begin{document}
\maketitle

\begin{abstract}
We propose \WS{}, a benchmark designed to assess the extent to which LLMs are consistently able to sustain tacit world models, by testing how they draw simple  inferences from descriptions of simple arrangements of entities. 
\WS{} is a synthetic benchmark with three problem types, each with their own trivial control, which explicitly avoids bias by decorrelating the abstract structure of problems from the vocabulary and expressions, and by decorrelating all problem subparts with the correct response. 
We run our benchmark on three state-of-the-art chat-LLMs (GPT3.5, GPT4 and Llama2-chat) and show that these models make errors even with as few as three objects.
Furthermore, they have quite heavy response biases, preferring certain responses irrespective of the question. 
Errors persist even with chain-of-thought prompting and in-context-learning.
Lastly, we show that while finetuning on similar problems does result in substantial improvements -- within- and out-of-distribution -- the finetuned models do not generalise beyond a constraint problem space.
\end{abstract}

\section{Introduction}

Large language models (LLMs)
display remarkable feats, such as the ability to perform new tasks given a small number of examples provided in context, or a verbal description of such tasks \citep{brown2020language}, or perform multi-step reasoning \citep{chowdhury-etal-2022-machine}.
Some of these abilities imply that LLMs have some form of grounded understanding of the outside world
\citep{bowman2023eight,gurnee2023language,roberts2023gpt4geo}. 
This is surprising, given that they have only been trained on text. 
For various reasons, however, it is challenging to evaluate how far LLMs' world understanding reaches and to what extent they create consistent internal representations that reflect the scenes in the world their input text describes.
Among these difficulties are the fact that benchmarks that test world understanding often have biases \citep[][i.a.]{gururangan-etal-2018-annotation,kavumba-etal-2022-prompt} and may be solvable through memorisation rather than understanding \citep[e.g.][]{sainz2023nlp,zhou2023don}.
Furthermore, a challenge is posed by the fact that world-understanding is inherently non-verbal -- referring to continuous states in a multidimensional world -- whereas language is strictly sequential and symbolic.
When models fail, it is thus difficult to say whether they lack an internal world model or simply fail to ``translate'' the language they receive to a world state, and back to language.

\begin{figure*}[h!]
	\centering
	\includegraphics[width=0.99\textwidth]{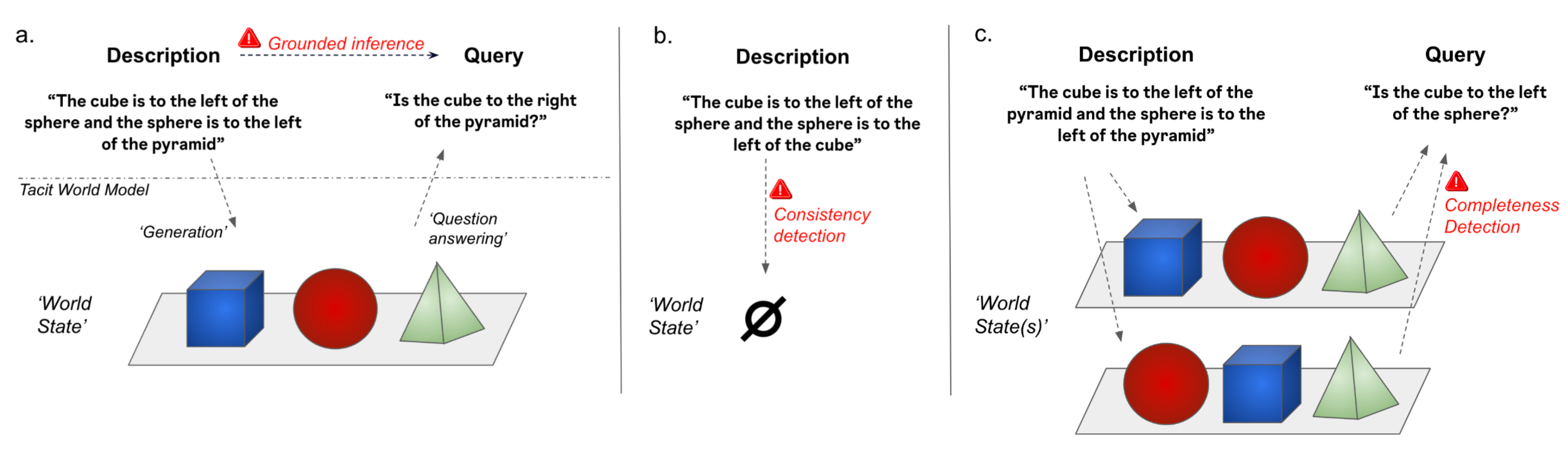}
	\caption{\textbf{The three problem types of \WS{}}. a. \textit{Grounded inferences} test a model's ability to generate a world state from a verbal description and inspect the world states to answer queries about them. For a language model, difficulties may arise in translating from text to world states or back, or in maintaining coherent world states. b. \textit{Consistency Detection} consists in detecting whether the verbal description contains a contradiction, and no possible world state can be generated from it. c. \textit{Completeness Detection} consists in detecting whether the description enables several alternative world states giving rise to conflicting responses to the query. }\label{fig:worldsenseproblems}
\end{figure*}

In this paper, we introduce \textit{\WS{}}, a benchmark inspired by tests in animals and young children, that tests for the existence of tacit world models in LLMs and addresses these challenges.
It escapes the limitations of previous benchmarks regarding bias and memorisation and pays attention to the specificities of the language-world translation problem. 
We focus on a class of problem where real world entities can be arranged in a linear order, and where the task is to draw transitive inferences based on a possibly partial description of the relative position of the entities. 
We chose this class of problem because it appears cognitively simple, transitive inference being readily drawn by humans adults \cite{lazareva2010nonverbal}, young children \cite{bryant1971transitive}, primates \cite{gillan1981reasoning}, rodents \cite{roberts1994transitive}, birds \cite{von1991transitive} and fishes \cite{grosenick2007fish}. 
In addition to that, linear order relationships are prevalent in language, and appear in several domains: spatial (``\emph{left}''\,/\,``\emph{right}'', ``\emph{above}''\,/\,``\emph{below}'', etc), temporal (``\emph{before}''\,/\,``\emph{after}''), and scalar (``\emph{more}''\,/\,``\emph{less}''). 
We therefore expect that language models should have ample opportunities to learn about both linear order and transitive inference from text alone. 

Concretely speaking, our benchmark contains three types of problems. 
The first type tests a model's ability to draw what we call \textit{grounded inferences} (see \cref{fig:worldsenseproblems}a). 
It corresponds to the ability to generate a representation of a world state from a verbal description, and to answer a verbal question about it. 
The second type, \textit{consistency detection}  (see \cref{fig:worldsenseproblems}b), tests the model's capability to spot whether descriptions describe possible scenes (i.e.\ correspond to a world state) versus impossible scenes (not corresponding to any world state, because of an internal contradiction). 
The third type, \textit{completeness detection} (see \cref{fig:worldsenseproblems}c), tests {if} a model is able to detect whether a description is complete or not. 
A complete description fully specifies a world state, whereas an incomplete description is compatible with several world states, and therefore may not allow to answer a given question. 

To ensure \WS{} is free of bias, we generate abstract program schemata, that we render into words in a variety of domains, ensuring that the responses are balanced and statistically independent from the particular choice of words in the problems.
To mitigate the world-language translation problem, we design `trivial' controls that separate the ability to create a coherent representation of a described scene from the ability to perform the reasoning steps on the input text required to understand the scene. 
Furthermore, our consistency and completeness test consider `epistemic' edge cases where the bidirectional translation between a verbal description and a world state fails because the description is incoherent or incomplete.

We apply our benchmark to three state-of-the-art chat-LLMs (GPT3.5, GPT4, and Llama2-chat) and find models can perform limited grounded inference for simple problems (with larger models being more capable).
However, we find models struggle with consistency and completeness.
Virtually all models are substantially biased in their responses.
Chain-of-thought prompting can help improve performance and -- for some models -- response bias, but only to a very limited extent and not across-the-board.
Furthermore, we find that when Llama2 models are finetuned explicitly on a training set containing novel instances of our problems, the resulting finetuned models generalise to new examples both in- and out-of-domain, but this generalisation is restricted to states involving linear relationships only, and does not improve reasoning or grounding beyond this specific set of states.
At the same time, models exhibit little evidence of memorisation of the data they were trained on, which we take as a promising sign for the contamination-sensitivity of our benchmark.
We release our \WS{} benchmark\footnote{\opensourceurl{}} to foster research into LLMs' capacity for building a tacit world model.

\section{Related Work}\label{sec:related_work}

In this section, we review several previous works that evaluate concepts related to \WS{}, such as common-sense understanding, more broad forms of reasoning and representational consistency.

\subsection{Commonsense understanding, world knowledge and reading comprehension}
A first line of work related to ours aims to assess the extent to which LLMs display world knowledge and commonsense understanding.
Most of those target commonsense understanding by formulating simple problems that can be resolved with world understanding -- e.g.\ ``\textit{How to separate egg whites from yolks using a water bottle?}''.
Examples of such benchmarks are HellaSwag, PiQA, SiQA, CommonsenseQA, ARC and GRASP \citep[respectively]{zellers-etal-2019-hellaswag,bisk2020reasoning,sap-etal-2019-social,talmor-etal-2019-commonsenseqa,clark2018think,jassim2023grasp}.
As these benchmarks provide scores that combine world knowledge (what are egg whites and yolks) and common-sense reasoning (how might you separate them with a water bottle), they do not paint a clear picture on either of these skills individually.
Some benchmarks combine questions with passages that consider relevant information \citep[e.g.\ OpenBookQA, TriviaQA;][]{mihaylov-etal-2018-suit,joshi-etal-2017-triviaqa}, and thus provide a somewhat clearer separation between these two facets, albeit at the cost of adding a third factor: reading comprehension.
While this alleviates the problem somewhat, such benchmarks also exemplify the issue outlined above, in cases where models obtain excellent scores \emph{without} having access to the accompanying evidence passage \citep[e.g.][]{touvron2023llama}.
Furthermore, it is generally unclear how much reported scores are impacted by dataset contamination.
\citet{touvron2023llama}, for instance, report that for several datasets -- including HellaSwag -- there is a strong correlation between contamination levels and sample scores.

\subsection{Mathematical reasoning}
A second line of work, in which reasoning and world knowledge are separated more strictly and which is related more closely to ours, investigates \emph{math world problems} (MVP) or mathematical reasoning more generally. 
There are various datasets available to do so, such as GSM8K \citep{cobbe2021training}, MMLU mathematics \citep{hendrycks2020measuring}, SVAMP \citep{patel-etal-2021-nlp} and the MATH dataset \citep{hendrycks2021measuring}, examples of which are all drawn from exams used to test mathematical knowledge and reasoning in humans at various levels.
For GSM8K and SVAMP, recent accuracies have been in the 90s for some models using specialised prompting techniches, whereas accuracies for MMLU mathematics and MATH are still well below  100.\footnote{Somewhat relatedly, there are several benchmarks that focus on LLMs ability to process \emph{code}, such as MBPP \citep{austin2021program} and HumanEval \citep{chen2021evaluating}. While one could argue that this requires reasoning as well, the focus of these benchmarks is more on the skill of programming than assessing reasoning.} 
Some studies put into question the extent to which these datasets really probe the  reasoning abilities of models.
Among other things, they raise concerns about bias in the test data \citep{yang-etal-2022-unbiased}, the impact of data leakage \citep{razeghi-etal-2022-impact}, the fact that the addition of irrelevant content\footnote{E.g.\ The addition of \emph{``Liz has 5 peaches''}, to the question ``\emph{Elsa has 5 apples. Anna has two more apples than Elsa. How many apples do they have together?}''} results in substantial performance drops \citep{shi2023large} and the robustness of reported results more generally \citep{stolfo-etal-2023-causal}.
In our benchmark, we try to avoid such issues by generating data not available online, from scratch, carefully controlling bias and spurious correlations, and adding several different controls for each problem.

\subsection{Separating form from meaning}
Many of the previously reported issues with datasets are rooted in the same issue: when evaluating `understanding' in LLMs, it proves difficult to separate their ability to output correct \emph{forms} from their ability to \emph{understand} those forms \citep[e.g.][]{ohmer2023evaluating,mitchell_krakauer_2023}.
Several studies have, in fact, shown that representations of LLM are not consistent across different lexical reprentations \citep[e.g.][]{ohmer2023evaluating,weber-etal-2023-mind}.
Here, we follow previous work in generating synthetic data that is unlikely to bear similarities to the training data in its specific form \citep[e.g.][]{weston-etal-2016-towards,gulordava-etal-2018-colorless}.

\section{\WS{}: grounded inferences, consistency, completeness}\label{sec:benchmark}

Having discussed the aims of our work and its positioning with respect to earlier studies, we now proceed to describe our benchmark, which we dub \emph{\WS{}}.
We describe how the benchmark's test set is algorithmically constructed using abstract schemata and textual skins to ensure absence of bias (\cref{subsec:algorithmic_generation}), provide information on the textual skins we use (\cref{subsec:skins});  we lay out the three problem types \WS{} contains -- grounded infererences, consistency detection and completeness detection (\cref{subsec:problem_types}), and we provide details about the generation of the \WS{} test set (\cref{subsec:testset}).

\begin{table*}
\caption{\textbf{Representative samples for the three problem types of \WS{} with three objects.}}\label{tab:experiment_examples}

\renewcommand{\arraystretch}{1.5}
\small
\begin{tabular}{|m{0.03\textwidth}|p{0.08\textwidth}<{\centering}|p{0.47\textwidth}|p{0.3\textwidth}<{\centering}|}
\hline
\textit{Pb.} & \textit{Condition} & \multicolumn{1}{c|}{\textit{Description}} & \textit{Query} \\
\hline
\multirow{3}{*}{\hspace{1mm}\rotatebox{90}{\small\it Inference \hspace{1mm}}} &
Normal  &   There are 3 objects arranged in a line: the bag is to the right of the box and the table is to the left of the box.
    & \multirow{2}{0.3\textwidth}[-2ex]{Is the following sentence `the table is to the left of the bag' TRUE or FALSE ? Only respond TRUE or FALSE without any explanation. }\\
    & Trivial & There are 3 objects arranged in a line. We list them from left to right: the table, the box, the bag.  &  \\
    \hline
\multirow{3}{*}{\hspace{1mm}\rotatebox{90}{\small\it Consistency \hspace{4.5mm}}} &
    Normal  & There are 3 objects arranged in a line. Someone says `the table is to the left of the bag, the table is to the left of the box and the bag is to the right of the box.'  & \multirow{2}{0.3\textwidth}[-2ex]{Is the situation just described possible or impossible ? Only respond with `POSSIBLE' or `IMPOSSIBLE' without any explanation.} \\
  & Trivial & There are 3 objects arranged in a line.
Someone says: `the book is to the left of the chair, the book is to the right of the mug and the mug is to the left of the chair.'&  \\ \hline
\multirow{3}{*}{\hspace{1mm}\rotatebox{90}{\small\it Completeness \hspace{0.8mm}}} &
   Normal  &There are 3 objects arranged in a line: the ball is to the left of the mug and the chair is to the right of the ball.  &\multirow{2}{0.3\textwidth}[-0ex]{(1) the chair is to the right of the mug; (2) the chair is to the left of the mug; (3) it is impossible to decide. Only respond with one of these options: (1), (2), or (3).}\\
 & Trivial &There are 3 objects arranged in a line: the ball is in front of the mug and the chair is in front of the ball.  & \\[4ex]
\hline
\end{tabular}
\end{table*}

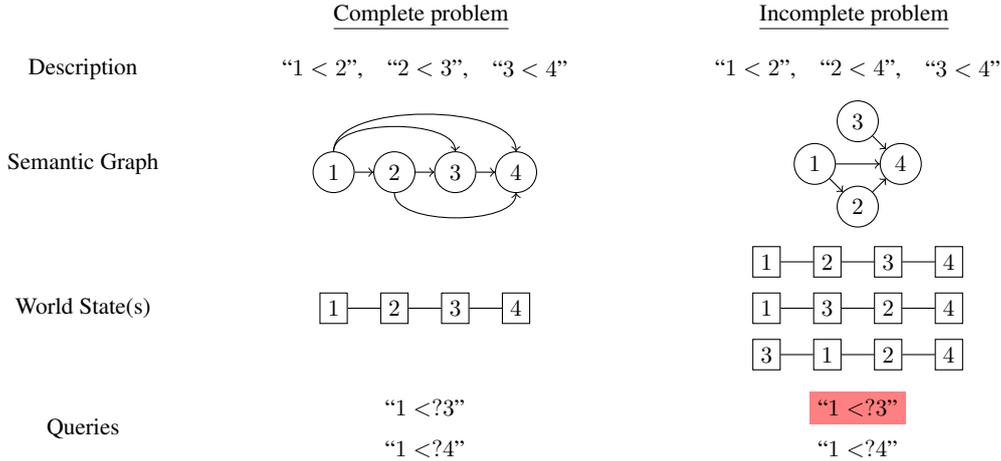
\begin{figure*}[h]
\centering
\begin{minipage}{0.2\textwidth}
\centering
\begin{tikzpicture}[main/.style = {scale=0.8}] 
\end{tikzpicture} 
\end{minipage}
\begin{minipage}{0.35\textwidth}
\centering
\begin{tikzpicture}[main/.style = {scale=0.8}] 
\node[main] (1) {\underline{Complete problem}}; 
\end{tikzpicture} 
\end{minipage}
\begin{minipage}{0.35\textwidth}
\centering
\begin{tikzpicture}[main/.style = {scale=0.8}] 
\node[main] (1) {\underline{Incomplete problem}};  
\end{tikzpicture} 
\end{minipage}
\vspace{4pt}

\begin{minipage}{0.2\textwidth}
\centering
\begin{tikzpicture}[main/.style = {scale=0.8}] 
\node[main] (1) {Description}; 
\end{tikzpicture} 
\end{minipage}
\begin{minipage}{0.35\textwidth}
\centering
\begin{tikzpicture}[main/.style = {scale=0.8}] 
\node[main] (1) {``$1 < 2$'',}; 
\node[main] (2) [right=0.1cm of 1]{``$2 < 3$'',}; 
\node[main] (3) [right=0.1cm of 2]{``$3 < 4$''}; 
\end{tikzpicture} 
\end{minipage}
\begin{minipage}{0.35\textwidth}
\centering
\begin{tikzpicture}[main/.style = {scale=0.8}] 
\node[main] (1) {``$1 < 2$'',}; 
\node[main] (2) [right=0.1cm of 1]{``$2 < 4$'',}; 
\node[main] (3) [right=0.1cm of 2]{``$3 < 4$''}; 
\end{tikzpicture} 
\end{minipage}

\vspace{4pt}
\begin{minipage}{0.2\textwidth}
\centering
\begin{tikzpicture}[main/.style = {scale=0.8}] 
\node[main] (1) {Semantic Graph}; 
\end{tikzpicture} 
\end{minipage}
\begin{minipage}{0.35\textwidth}
\centering
\begin{tikzpicture}[main/.style = {draw, circle,scale=0.8}] 
\node[main] (1) {$1$}; 
\node[main] (2) [right of=1]{$2$}; 
\node[main] (3) [right of=2]{$3$}; 
\node[main] (4) [right of=3]{$4$}; 
\draw[->] (1) -- (2);
\draw[->] (2) -- (3);
\draw[->] (3) -- (4);
\draw[->] (1) to [out=90,in=90,looseness=.7] (3);
\draw[->] (1) to [out=90,in=90,looseness=.7] (4);
\draw[->] (2) to [out=270,in=270,looseness=.7] (4);
\end{tikzpicture} 
\end{minipage}
\begin{minipage}{0.35\textwidth}
\centering
\begin{tikzpicture}[main/.style = {draw, circle,scale=0.8}] 
\node[main] (1) {$1$}; 
\node[main] (2) [below right of=1]{$2$}; 
\node[main] (3) [above right of=1]{$3$}; 
\node[main] (4) [below right of=3]{$4$}; 
\draw[->] (1) -- (2);
\draw[->] (3) -- (4);
\draw[->] (1) -- (4);
\draw[->] (2) --(4);
\end{tikzpicture} 
\end{minipage}

\vspace{6pt}
\begin{minipage}{0.2\textwidth}
\centering
\begin{tikzpicture}[main/.style = {scale=0.8}] 
\node[main] (1) {World State(s)}; 
\end{tikzpicture} 
\end{minipage}
\begin{minipage}{0.35\textwidth}
\centering
\begin{tikzpicture}[main/.style = {draw,scale=0.8}] 
\node[main] (1) {$1$}; 
\node[main] (2) [right of=1]{$2$}; 
\node[main] (3) [right of=2]{$3$}; 
\node[main] (4) [right of=3]{$4$}; 
\draw (1) -- (2);\draw (2) -- (3);\draw (3) -- (4);
\end{tikzpicture} 
\end{minipage}
\begin{minipage}{0.35\textwidth}
\centering
\begin{tikzpicture}[main/.style = {draw,scale=0.8}] 
\node[main] (1) {$1$}; 
\node[main] (2) [right of=1]{$2$}; 
\node[main] (3) [right of=2]{$3$}; 
\node[main] (4) [right of=3]{$4$}; 
\node[main] (5) [below=0.2cm of 1]{$1$}; 
\node[main] (6) [right of=5]{$3$}; 
\node[main] (7) [right of=6]{$2$}; 
\node[main] (8) [right of=7]{$4$}; 
\node[main] (9) [below=0.2cm of 5]{$3$}; 
\node[main] (10) [right of=9]{$1$}; 
\node[main] (11) [right of=10]{$2$}; 
\node[main] (12) [right of=11]{$4$}; 
\draw (1) -- (2);\draw (2) -- (3);\draw (3) -- (4);
\draw (5) -- (6);\draw (6) -- (7);\draw (7) -- (8);
\draw (9) -- (10);\draw (10) -- (11);\draw (11) -- (12);
\end{tikzpicture} 
\end{minipage}

\vspace{7pt}
\begin{minipage}{0.2\textwidth}
\centering
\begin{tikzpicture}[main/.style = {scale=0.8}] 
\node[main] (1) {Queries}; 
\end{tikzpicture} 
\end{minipage}
\begin{minipage}{0.35\textwidth}
\centering
\begin{tikzpicture}[main/.style = {scale=0.8}] 
\node[main] (1) {``$1<?3$''}; 
\node[main] (2) [below=0.1cm of 1]{``$1<?4$''}; 
\end{tikzpicture} 
\end{minipage}
\begin{minipage}{0.35\textwidth}
\centering
\begin{tikzpicture}[main/.style = {scale=0.8}] 
\node[main] (1) [fill=red!50]{``$1<?3$''}; 
\node[main] (2) [below=0.1cm of 1]{``$1<?4$''}; 

\end{tikzpicture} 
\end{minipage}%

\caption{\textbf{Generation of \WS{} completeness problems.} Left: Complete problem with 4 entities. Right: Incomplete problem derived from the complete problem. Descriptions are list of verbalised binary or ternary relations. The semantic graph represent the underlying total or partial order. World state(s) are represented as objects in a 1D left-to-right disposition. Queries are verbalised relations, whose truth values are either determinate (white) or undeterminate (red). The incomplete problem is generated from the complete problem by first randomly picking a relation from the description say $``2<3"$; randomly pick either the left entity `2' or the right entity `3' and replace it with its direct neighbour resp. `1' or `4'. The new relation becomes resp. $``1<3"$ or $``2 < 4"$. }\label{fig:graph}
\end{figure*}

\subsection{Algorithmic problem generation}\label{subsec:algorithmic_generation}
Each problem in \WS{} is generated in an abstract format, containing a \textit{world state}, a \textit{description} and a \textit{query} (see also \cref{fig:graph}).

\paragraph{World state}
The world state can be viewed as a list of entity indices plus their position (or other scalar properties)  in the world (e.g.\ ($object_1$, $position_1$), ... ).

\paragraph{Description}
The description of the world state consists of a list of binary or ternary relations between object indices (e.g.\ $object_1$ \texttt{\footnotesize to\_the\_left\_of} $object_2$, or $object_1$ \texttt{\footnotesize is\_in\_between} $object_2$ and $object_3$).
A description typically specifies a single world state (when the description is complete), but it can also be compatible with more than one state (when it is incomplete), or none at all (when it is inconsistent). 

\paragraph{Query}
The third component of each problem is the question asked to the model about the description of the world state, which we call a \emph{query}.
In \emph{inference problems}, the query asks if a  specific binary or ternary relationship is true.
\emph{Consistency problems} have no relationships in the query, but simply ask if the entire description is possible or impossible.
For \emph{completeness problems}, the query contains three options: a relationship, its opposite and `impossible to decide'.\footnote{
More information on how exactly the various problems are generated can be found in \cref{subsec:problem_types}.}

\subsection{The \WS{} skins and problem rendering}\label{subsec:skins}

Once generated, descriptions and queries are rendered into text using \textit{skins}, which turn entities into noun phrases (e.g.\ ``\textit{the red ball}''), and relations into verbal descriptions (``\textit{to the left of}''). 
The descriptions and queries are assembled into a problem instance presented to the {L}LM. 
Each problem instance comes with a set of acceptable responses (e.g.\ \textttt{TRUE} or \textttt{FALSE}), 
which are verbally presented to the {L}LM as a forced choice (e.g.\ \textit{``Only respond TRUE or FALSE without any explanation"}).
In the present \WS{} benchmark, we restrict world states to be of the simplest possible kind: a finite set of $N$ entities arranged according to a total order. 
We use skins that belong to three domains: spatial (left versus right, etc.), temporal (before versus after) and scalar (larger versus smaller, etc). 
Each domain contains six skins and has entities that belong to various types  (objects, people's names, events, places, etc).
For descriptions and queries, we use two opposite binary order relations (e.g.\ less-than, more-than) and one ternary relation (e.g.\ in-between). 
The queries are testing the ability of models to perform transitive inferences (if $a<b$ and $b<c$, then $a<c$). 
A full description can be found in \cref{tab:supp-scenes}.\footnote{\scriptsize \url{http://github.com/facebookresearch/worldsense}}

\subsection{Problem types}\label{subsec:problem_types}

\WS{} contains three types of problems: grounded inference problems, consistency problems, and completeness problems. 
Below, we describe the problem types in more detail; examples can be found in \cref{tab:experiment_examples}.

\paragraph{Grounded inferences}
Grounded inference problems assess a models' ability to answer questions based on textual descriptions of simple scenes with entities. 
This corresponds to two abilities bundled together: the ability to generate a representation of a world state given a description and the ability to answer questions about that world state.
To distinguish them, we include a `trivial' control experiment in which the description is replaced by an ordered list of entities which represents an iconic rendering of the underlying world state. 
In this condition, the verbal description is itself homothetic with the world states, making the generation of a tacit world state trivial or not necessary. 
The `only' difficulty is answering the question by reading out the answer from a provided world state. 
In the normal condition, all descriptions are complete. 
The relations in the description refer to adjacent entities in the underlying world state, but by contrast with the \emph{trivial control}, they are given in a scrambled order during verbal rendering. 
The query contains either a binary relation or a ternary relation which was not part of the description and therefore refers to non-adjacent entities. 

\paragraph{Consistency problems}
Consistency problems focus on the epistemic distinction between a verbal description and a world state. 
More specifically, we investigate whether LLMs can detect if descriptions are \emph{inconsistent}, i.e.\ contain a contradiction, and hence compatible with no world state.
To generate such problems, we start from the inference problem set, and we modify it by incorporating what was the query in a random position in the description,thus turning it into descriptions that are contradictory (if the query was false) or redundant (if the query was true). 
We then ask the model whether that description is possible or impossible. 
As above, the inserted query can be binary or ternary.
In the trivial control condition of this experiment, we do not use the question query from a grounded inference problem, but instead one of the relations of the description itself, and either contradict it (impossible) or repeat it (possible).
In this control, the task of spotting contradictions becomes much simpler and can be done by inspecting the text only, without necessarily generating world states. 

\paragraph{Completeness problems}\label{subsec:completeness}
Lastly, completeness problems evaluate another epistemic distinction: that between \emph{complete} and \emph{incomplete} descriptions. 
A complete description fully specifies a world state, whereas an incomplete description is compatible with several world states. 
For this problem type, we build a new set of problems, changing the underlying world states to be a partial order. 
For an example of this process, we refer to \cref{fig:graph}.
Once the partial orders are generated, we build three possible queries: one that is true, one that is false, and one that is indeterminate.
For example, given the description ``\textit{Paul is taller than Jack and Paul is taller than Marc.}'', we render the query as 3 alternative forced choices: ``(1) \textit{Jack is taller than Marc} (2) \textit{Jack is shorter than Marc} (3) \textit{it is impossible to decide}''. 
In the trivial control conditions the indeterminate questions are obtained by either substituting one of the entities by a new entity not present in the description e.g ``\textit{Paul is taller than Jack and Jack is taller than Marc. Is Alfred taller than Marc?''} or a new relation not introduced before ``\textit{Paul is taller than Jack, Jack is taller than Marc. Is Paul to the left of Marc?''}. 
In the trivial condition, detecting incompleteness can be done simply by scanning the text for new entities or relations only mentioned in the query and does not necessitate building a world model.
As shown in Table \ref{tab:experiment_examples}, the trivial problems are presented as a 3 alternative forced choices.

\subsection{The \WS{} test set}\label{subsec:testset}
To generate the full \WS{} test set, we first generate each problem as a tuple $t$, where the description is kept constant and only the query varies, to yield all of the possible truth values. 
This ensures that over the test set all truth values are equiprobable, and it helps reduce the variance in the analysis, since accuracies and response biases can be computed for each tuple before being aggregated by condition. 
For the inference and consistency problems $|t|=2$, for the completeness problems $|t|=3$. 
Furthermore, we generate the world states, descriptions and queries of the test set such that the ground truth values are statistically independent from the description and query, both in the abstract and when rendered as text.
This guarantees that only looking at the description or query in isolation to solve the problem will yield chance level performance. It rules out obtaining good performance via low level shortcuts.\footnote{In fact, in our problems, the ground truth is statistically independent from any n-grams of relations extracted from the problem statement (where $n$ is strictly less than the total number of relations).}
We generated a total of around 90K problems, spread out roughly equally across the three problem types, with equal representations for each domain, skin and size.
The complete test set can be found in the \texttt{data} folder of our github repository.\footnote{\opensourceurl{}}

\section{Experimental setup}\label{sec:setup}

To showcase and test our benchmark, we evaluate three chat-LLMs: OpenAI's Chat GPT3.5,  GPT4 and a Llama2-chat model based on Meta's Llama2 LLM. 
Here, we report details on how we run them as well as how we compute both accuracy and response bias for their responses.

\subsection{Models}

For all models, we feed the prompt to the model as if typed by a user and then recover the generated text. 
Each problem instance is treated as a new conversation. 
To run OpenAI's models, we use the models \texttt{gpt-3.5-turbo} and \texttt{gpt-4} from the OpenAI API.
We use the default parameters of the interface for chat generation; (the experiments were run between August and October 2023). 
For Llama2-chat, the chat version of Llama2 70B, we use the HuggingFace model \textttt{Llama-2-70b-chat-hf}), on which we run inference directly, using greedy decoding with a temperature of 0. 

\subsection{Evaluation}
To evaluate the response of the model, we select the last word of the generated text and normalise it by upcasing and removing extra punctuation. 
If this does not match the set of acceptable answers as specified in the prompt (e.g.\ \textttt{TRUE} or \textttt{FALSE} for inference problems), the model receives a continued  conversation where the user asks: ``\textit{What is the final answer? 
Respond only using one of these possible answers: TRUE, FALSE}''.  
If the final answer correspond to the ground truth, it is deemed correct, and incorrect otherwise.

\paragraph{Accuracy}
The main metric we use for \WS{} is accuracy.
We compute accuracy on a tuple-by-tuple basis (where each tuple corresponds to minimally different problems, one for each possible ground-truth answer, see \cref{subsec:testset}).
For each trial in the tuple, we set the trial's accuracy to $1$ if the response matches the ground truth, $0$ otherwise and then average this score within the tuple. 
For the completeness problems, we fold the three-way forced choice into a binary choice (responses 1 and 2 are folded as \texttt{KNOWN}, and response 3 as \textttt{UNKNOWN}). 
To obtain a balanced score, we compute a weighted average where the trials corresponding to \textttt{KNOW} are given a weight of 0.5 and the trials corresponding to \textttt{UNKNOWN} a weight of 1. 
After such averaging and rebalancing, all of the accuracies across tuples and problem types have a chance level of 0.5.  

\paragraph{Response bias}
Earlier in this work, we noted that previous benchmarks are often not balanced across responses.
In our benchmark, we ensure that responses are balanced across different dimensions (condition, skin, domain, problem type, etc).
This ensures that models cannot benefit from biased responses and makes it less likely that there are unintentional spurious correlations in the problem set that could be exploited.
It also implies that we can meaningfully study a model's \emph{response bias}.
We compute the response bias by assigning all positive polarity responses (\texttt{TRUE}, \texttt{POSSIBLE} and \texttt{KNOWN}) a score of $1$, and all negative responses (\texttt{FALSE}, \texttt{IMPOSSIBLE} and \texttt{UNKNOWN}) a score of $-1$.
Following the computation of accuracy, the response bias for a particular group is then defined as the average -- weighted, if there are more than 2 responses -- between all bias scores in that group.
The resulting bias score is bound between 1 and -1, where 0 implies absense of bias, and 1 and -1 indicate a systematic bias to positive or negative responses, respectively.


\section{Results}\label{sec:results}
We now report the results of our experiments.
First, we discuss general benchmark scores for the three models under consideration (\cref{subsec:benchmark_scores}) as well as trends in their response bias (\cref{subsec:response_bias}).
Then, we explore several methods to improve those scores, such as employing in-context-learning (ICL) or chain-of-thought (COT) prompting (\cref{subsec:prompting}) and finetuning on similar data (\cref{subsec:finetuning}).

\subsection{Benchmark scores}\label{subsec:benchmark_scores}

\begin{table}[]
\caption{\textbf{\WS{} main result.} Average accuracies by 3 chat-LLMs and a Llama2$_{70B}$ model finetuned with instruction-tuning data plus 1M training samples, which we dub Llama2-FT+1M, with 95\% confidence intervals. Chance level is 50\%.}\label{tab:basic_results}

\setlength{\tabcolsep}{2pt}
\centering
\begin{tabular}{lcc c  c   c}
\toprule
& \multirow{2}{*}{GPT3.5} & \multirow{2}{*}{GPT4} && \multicolumn{2}{c}{Llama2} \\
\cline{5-6}
&           &         &&  -chat                  & -FT+1M\\
\hline
WS avg. & 55.6 {\tiny $\pm 0.4$} & 75.6 {\tiny $\pm 0.4$} && 56.2 {\tiny $\pm 0.3$} & 77.4 {\tiny $\pm 0.4$} \\
\bottomrule
\end{tabular}
\end{table}

\begin{figure*}
\begin{subfigure}{0.5\textwidth}
\includegraphics[width=0.99\columnwidth]{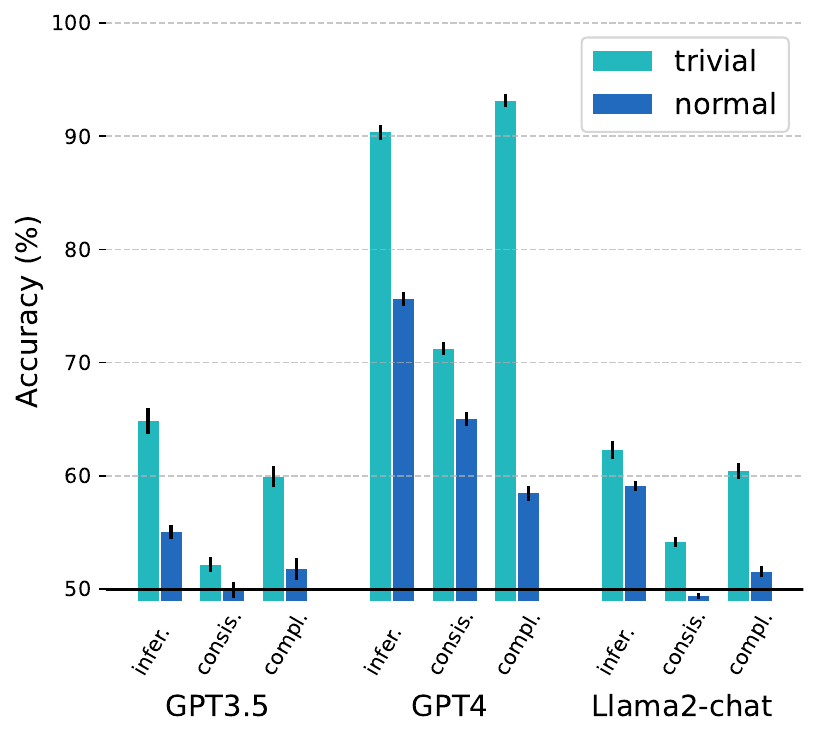}
\label{fig:exp1_accuracy}
\end{subfigure}
\begin{subfigure}{0.5\textwidth}
\includegraphics[width=0.99\columnwidth]{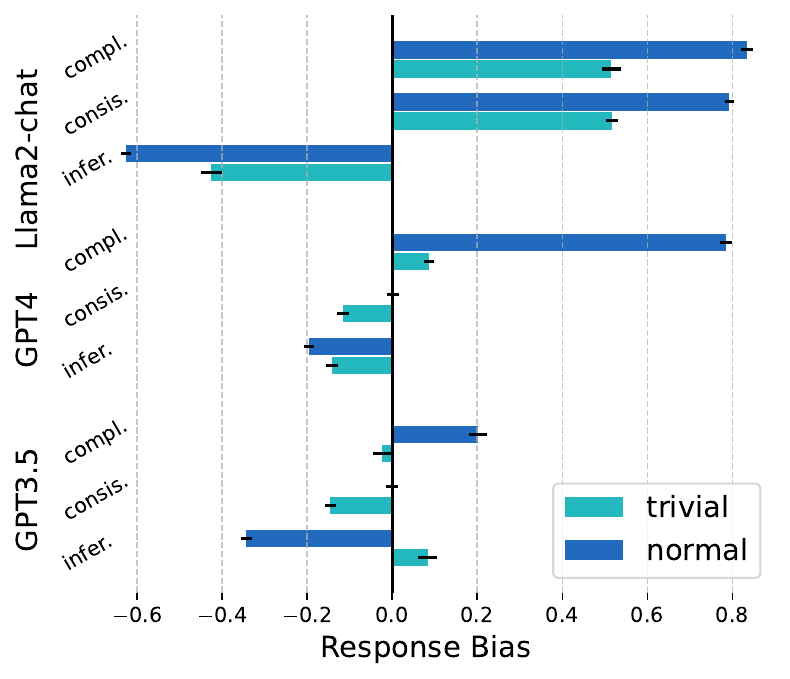}
\label{fig:exp1_response_bias}
\end{subfigure}
    \caption{\textbf{Main \WS{} results across problem types.} Left: Accuracy of three chat models split out by problem type (inference, consistency, completeness) and condition (trivial, normal). The horizontal black line indicates chance level. Right: Response bias of models split out by problem type and condition. +1 indicates a positive bias (\textttt{TRUE}, \textttt{POSSIBLE}, \textttt{KNOWN}), -1 a negative bias (\textttt{FALSE}, \textttt{IMPOSSIBLE}, \textttt{UNKNOWN}), 0 indicates no response bias. For both plots, error bars denote 95\% confidence intervals.}\label{fig:exp1}
\end{figure*}

In \cref{tab:basic_results}, we report the average benchmark scores of GPT3.5, GPT4 and LLaMA2-chat on \WS{}.
We see that GPT4 outperforms the other two models by quite a large margin.
GPT3.5 and Llama2-chat are barely above chance levels.
In \cref{fig:exp1} (left), we visualise the results in more detail, split out by problem type (inference, consistency and completeness) and condition (normal and trivial).
This figure illustrates that the `epistemic' problems (consistency and completeness) yield systematically lower performances than the grounded inferences, across all models.
Furthermore, the difference between the trivial control and the base experiment shows that -- at least for GPT4 -- a large fraction of the error is due to (tacit) world state generation, while the other models also have problems with the question-answering component of the benchmark problems.
In the Appendix, we provide the corresponding numerical results (\cref{tab:supp_exp1}), and a break down by number of objects (problem size) and complexity (\cref{fig:exp1_by_sizeco}, \cref{tab:sup_monster_acc}).
Unsurprisingly, models make more errors when there are more entities.
Interestingly, this is true both for the trivial and the normal case.
Furthermore, all models struggle with ternary queries.
Lastly, we see that some problem domains seem  more difficult than others: for all models scalar problems are the easiest, while the temporal and spatial problems are more challenging (\cref{tab:exp1_by_domain}).

\subsection{Response bias}\label{subsec:response_bias}
Next, we study the models' response bias across condition and problem type, which we plot in \cref{fig:exp1} (right).
We see that especially Llama2-chat is quite heavily biased in its responses, preferring to respond \texttt{FALSE} more than 80\% of the times for the inference problems, and strongly preferring to respond \texttt{POSSIBLE} and \texttt{KNOWN} for the consistency and completeness problems. 
GPT4's and GPT3.5's response biases are much smaller, except for the normal completeness and inference problems, where GPT4 has a strong KNOW, and GPT3.5 a FALSE bias, respectively.

\subsection{Impact of prompting strategies}\label{subsec:prompting}
In our initial experiments, we presented models directly with a description and corresponding question.
From these experiments, we concluded that none of the three models shows clear signs of creating tacit world models.
Next, we investigate whether more promising conclusions 
could be drawn if we use more sophisticated prompting strategies, known to improve the reasoning abilities of LLMs.
To this end, we evaluate models both in an ICL (few-shot, in-context learning) setup, in which we prepend five examples together with their solution to the models' input, and with Chain-Of-Thought (COT) prompting, in which we ask the model to break down a complex problem into smaller ones.
For the ICL setup, we consider two different conditions: in the within-distribution-condition (IID), the examples have the same problem size, complexity, skin and domain as the test problems; in the out-of-domain (OOD) setup, the sample problems have the same number of entities and complexity, but a different skin and domain.
For the COT setup, we provide a detailed instruction to the model, where we ask it to devise a strategy and then execute it. 
We provide example of the COT prompt in \cref{app:COT_prompts} and ICL example in
\cref{subsec:ICL_example}.

\begin{figure*}
\begin{subfigure}{0.5\textwidth}
\centering
\includegraphics[width=0.99\columnwidth]{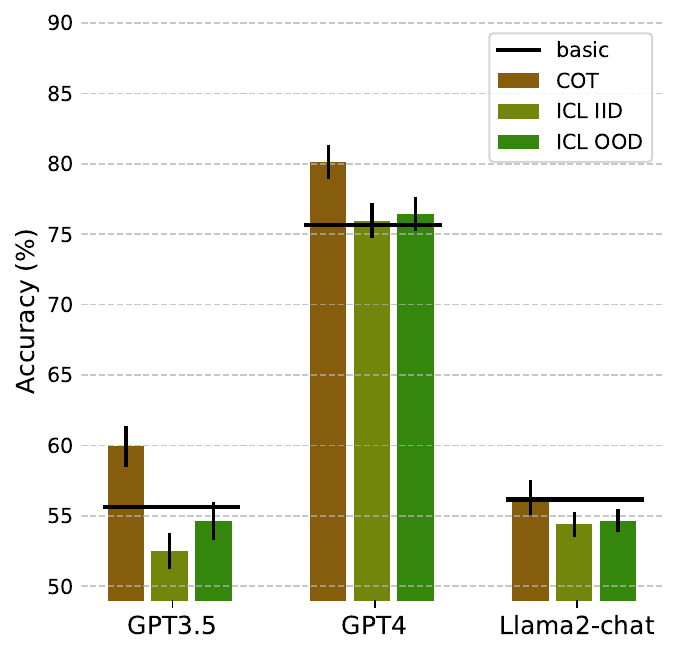}
\label{fig:exp2_accuracy}
\end{subfigure}
\begin{subfigure}{0.5\textwidth}
\includegraphics[width=0.99\columnwidth]{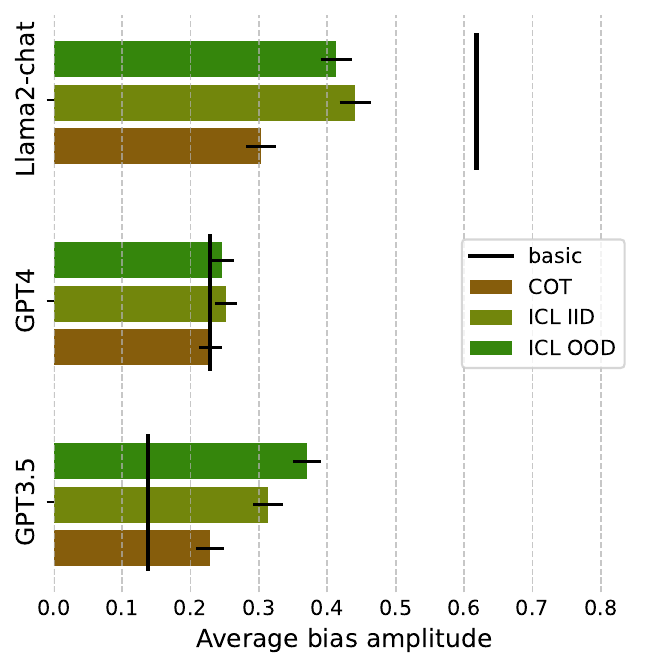}
\label{fig:exp2_bias}
\end{subfigure}
\caption{\textbf{Prompting enhancement results.} Left: Accuracy across different prompting strategies, averaged over problem types and conditions. Right: Average response bias amplitude (absolute value) across the 6 problem types.  
The black line indicates the values corresponding to the basic prompting setup. Chance levels for accuracies are 50\%, error bars denote 95\% confidence intervals for both plots.}\label{fig:exp2}
\end{figure*}

In \cref{fig:exp2}, we plot, under these different prompting strategies, the models' accuracies (left) and response bias \emph{amplitudes} (right) averaged across all problem types (see Table \ref{tab:supp_exp2} for a breakdown by problem type).
In terms of accuracies, COT prompting provides a slight benefit for both GPT3.5 and GPT4 (4.3 and 4.6 percentage point, respectively).
For Llama2-chat, there appear to be no statistically significant changes in accuracy by applying any of these setups.
None of the models benefit statistically significantly from either of the ICL setups.
Interestingly, however, both setups result in a very substantial decrease in response bias for Llama2-chat, suggesting that these setups do change the actual responses given by that model.
For GPT3.5, intriguingly, using our alternative prompting strategies \emph{increases} the response bias.
For all models, COT prompting results in lower response biases than the ICL setups.
Taken together, these results strengthen the conclusions from the previous section: even with more sophisticated prompting strategies to extract information from the models, their \WS{} scores are far from perfect.

\subsection{Finetuning models}\label{subsec:finetuning}

Lastly, we test whether chat models can \emph{learn} to do grounded inferences when we explicitly finetune them on similar problems.
To generate a finetuning training set, we selected four scalar and four temporal skins from the \WS{} testset, which we populated with novel entities (see \cref{tab:finetuning_entities}).
To increase diversity, we also created five new scalar and five new temporal skins, which we also populated with novel entities.
We then built two training sets of increasing sizes: 100K samples and 1M samples, which we mixed into the standard 2B token finetuning training set for Llama2-chat, and we used them to finetune Llama2-70B.\footnote{We finetune the model for 200K iterations (amounting to approximately 2 epochs), with a learning rate of 2e-6 and a batch size of 64, using a warmup of 2000 steps and a cosine scheduler. We set the max sequence length to 4096.}
We also finetune a model on the standard trainingset without mixing in any \WS{} data.\footnote{Note that we do not apply the RLHF stage, so the model that is finetuned without \WS{} data is not identical to the model used in our first experiments.}

To test the finetuned models, we first test how they fare on the standard test set of \WS{}.
Because of how we constructed our training set, the \WS{} test set contains a mix of IID skins (the eight scalar and temporal skins that were present in the training data -- albeit with different entities and different versions) and OOD skins (the two remaining scalar and temporal skins, and the entire spatial domain).
We report results for these two subsets (\WS{} IID and \WS{} OOD) separately.
To test both memorisation and generalisation more elaborately, we also construct various other test sets.
First, we test memorisation computing scores for 10K examples that were in the training set (first column).
In addition to that, we generate a \emph{ood-size} set, containing problems of size six (finetuning set has sizes 3,4,5); an \emph{ood-query} set with situations described as before, but where we ask a new type of query never seen in the finetuning set, e.g. asking about first or last instead of before, after, between relations;
and \emph{ood-problem} a completely different set of problems which does not correspond to a linear order.
Examples can be found in \cref{tab:ood_examples}.
Lastly, we test our finetuned models on three standard benchmarks involving reasoning and/or world knowledge: MMLU, GSM8K and MATH \citep[respectively]{hendrycks2020measuring, cobbe2021training,hendrycks2021measuring}.

\begin{figure*}[]
\begin{subfigure}{0.5\textwidth}
\includegraphics[width=.99\columnwidth]{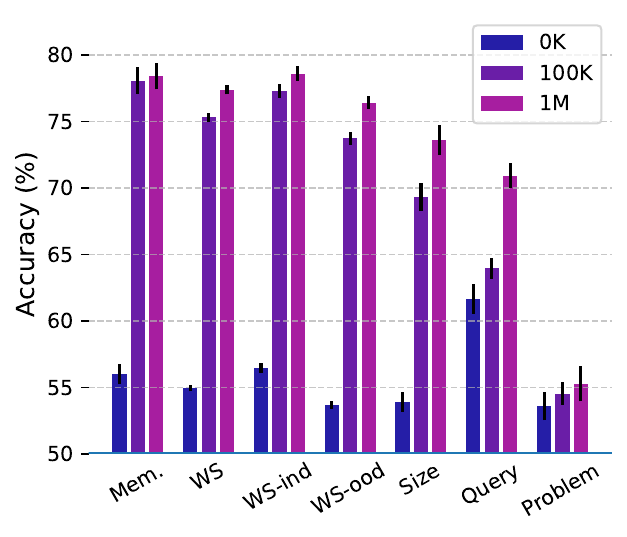}
\end{subfigure}
\begin{subfigure}{0.5\textwidth}
\includegraphics[width=.97\columnwidth]{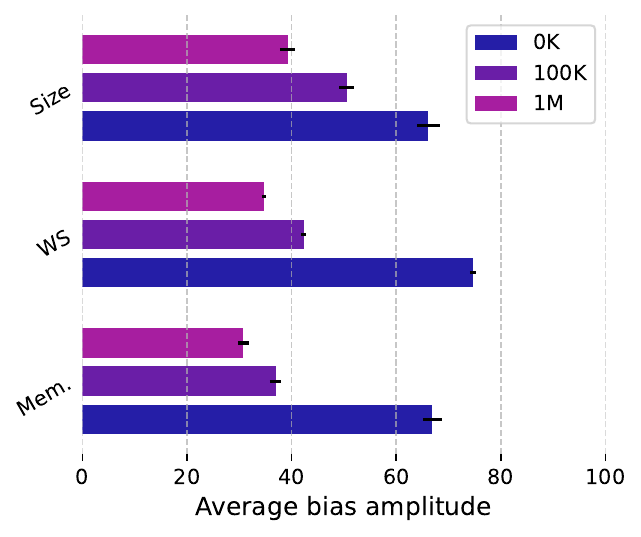}
\end{subfigure}
\caption{\textbf{Finetuning results.} Left: Accuracy across  Llama2$_{70B}$ models finetuned on 0, 100K and 1M training examples, on the \WS{} (WS) test set split into in-domain (WS-ind) and out-of-domain (WS-ood) subsets, plus a memorisation test set (Mem.), and the ood-size, ood-query and ood-problem generalisation test sets.  Right: Bias amplitude on the memorisation, WS and length generalisation test sets. Chance levels for accuracies are 50\%, error bars denote 95\% confidence intervals for both plots.}\label{fig:exp3_accuracy}
\end{figure*}

As can be seen in \cref{fig:exp3_accuracy}, the finetuned models improve their performance on \WS{}, both in-domain and out-of-domain (the best overall score is reported in \cref{tab:basic_results}).
At the same time, there is hardly any evidence of memorisation -- accuracies on the memorisation set are comparable with the overall \WS{} scores -- indicating that the benchmark is not particularly contamination sensitive. 
The various generalisation datasets show that what is learned during finetuning generalises to different entities, new relations and domains and similar problems with more entities. Note, though, that not all problem types are learned equally well, with consistency problems being the hardest to learn (Table~\ref{tab:supp_exp3}).
There is, however, not much evidence of learning beyond the linear topology of world-states that all those generalisation sets share: there are no substantial improvements for world states with non-linear orders, or for external reasoning / \WS{} benchmarks (see \cref{tab:exp3_external}).
Our finetuning protocol did thus not improve grounded reasoning and tacit world models beyond a small problem space.

\section{Discussion}\label{sec:conclusion}

In this work, we explore the ability of LLMs to verbally reason about spatial, temporal and scalar properties of objects or other entities as if they were able to represent these objects in an internal world model. 
To do so, we propose \WS{}, a benchmark that focuses on \textit{grounded transitive inference} problems, in which a model is asked to answer simple questions about verbal descriptions that describe world states in which up to five entities are arranged according to a sequential order.
Both our problem selection (transitive inferences for linear order relationships) and problem construction (systematically, and bias-free) are inspired by cognitive science.
We include three different problem types -- grounded inference, consistency detection and completeness detection -- and include trivial controls for all these three problems.

To test and showcase our benchmark, we evaluate three state-of-the-art chat-LLMs: GPT3.5, GPT4 and Llama2-chat.
We show that some models (in particular: GPT4) are able to answer questions with relatively few errors if the world state is simple and presented directly as a linear sequence (e.g.\ entity1, entity2, entity3), while other models (GPT3.5, LLama2-chat), make many errors even in that simplest scenario.
However, for all models that we tested, errors start to accrue massively when world states have to be constructed from a verbal description, indicating that models struggle to translate the language they receive into an internal world model.
In accordance with that, \WS{} illustrates the inability of models to grasp the fact that certain descriptions are inconsistent and correspond to no possible world state, or incomplete, making it impossible to answer certain questions.
Furthermore, all models have substantial response biases: they systematically prefer some answers over others, regardless of the problem specifics.
High error rates persist even with alternative prompting strategies, such as chain-of-thought prompting and in-context learning.

In a follow-up experiment, we investigate whether models can \emph{learn} to do grounded inferences when we explicitly finetune them on data similar to the \WS{} benchmark.
We find that a finetuned Llama2 model in fact improves substantially on both in-domain and out-of-domain problem sizes, entities and queries, while showing little evidence of memorising the data that it has been explicitly trained on.
Yet, these improvements appear restricted to the class of problems with transitive relationships: finetuning does not result in improvements on problems in which the relationships between entities are non-transitive (e.g. lines that are parallel or perpendicular), nor in improvements in performance on external reasoning benchmarks.
To allow further exploration, we release  our data, including our finetuning dataset and various within- and out-of-distribution test sets.

With \WS{}, we aim to provide a systematic and bias-free benchmark to test for the existence of tacit world-models in LLMs.
Such world models are useful in that they enable organisms to run a predictive model in the world state and guide their behavior based on these predictions. 
We restricted ourselves to the case where a verbal description corresponds to a static scene. 
As shown, models already make many errors on this account.
Yet, the benchmark only scratches the surface of the evaluation of world modelling and reasoning abilities of LLMs. 
In further iterations of \WS{}, we aim to address the ability of the models to follow dynamic changes in world states through the action of the laws of physics or agents, envision counterfactuals, and plan in this representation. 

\section{Acknowledgments}

The work reported here conducted by ED in his EHESS capacity was funded by the Agence Nationale pour la Recherche (ANR-17-EURE-0017 Frontcog, ANR-10-IDEX-0001-02 PSL*, ANR-19-P3IA-0001 PRAIRIE 3IA Institute) and a grant from CIFAR (Learning in Machines and Brains). We thank E. Chemla and the CoML team for useful comments.

\bibliography{anthology,custom}
\bibliographystyle{acl_natbib}

\appendix
\onecolumn
\section{Benchmark details}\label{appendix:benchmark_details}

The \WS{} benchmark has one primary test set, a finetuning train dataset, and various out-of-distribution test sets.
In this section, we provide details for all these different partitions.
The full datasets can be downloaded from our \href{https://github.com/facebookresearch/worldsense#test-sets}{github repository}.

\subsection{The \WS{} test set}
The standard \WS{} consists of three types of problems (inference, consistency, and completeness), each with a trivial control, spanning three domains (spatial, temporal, and scalar).
Spatial relationships, for example, test whether an entity is to the left or to the right of another entity, e.g.\ ``\textit{Is the red ball to the left of the blue box?}''.
Temporal relationships involve questions requiring a sense of time often formulated with before or after statements.
Scalar relationships describe whether one scalar is larger or smaller than another, an inference we often make in the form of say comparing one person's height to another's. 
In \cref{tab:supp-scenes}, we report the entities and relations used for the three different domains in the \WS{} test set.
The `skins' we use to transform the abstract schemata created with these entities and relations can be found in the main text of this paper, in \cref{tab:experiment_examples}.

\begin{table*}[!h]
    \caption{\textbf{Entities and relationships in \WS{} test set}. Some relations occur also in the finetuning dataset, which we denote with *.}\label{tab:supp-scenes}
    \scriptsize
    \renewcommand{\arraystretch}{1.5}
\begin{tabular}{|m{0.05\textwidth}<{\centering}|
m{0.43\textwidth}|
m{0.19\textwidth}|
m{0.22\textwidth}<{\centering}|}
\hline
{\textbf{Domain}} & \multicolumn{1}{c|}{\textbf{Entities}} & \multicolumn{1}{c|}{\textbf{Binary relations}} & \textbf{Ternary Relations} \\
\hline
\multirow{3}{*}{\rotatebox{90}{\small Spatial}} &
    \emph{Objects:} red ball, blue box, yellow chair, orange table, purple bag, brown mug, green guitar, white bicycle, black book, gray briefcase, pink shoe, silver watch &
    \emph{is} to the left of/right of, in front of/behind, above/below &
    \{name\} \emph{is} in between \{object\} and \{object\} \\ 
 &
    \emph{People}: James, David, Robert, Michael, William, Frank, Alice, Claire, Linda, Patricia, Emily, Grace &
    \emph{is} to the left of/right of, in front of/behind,
    \{name\} \emph{lives} above/below &
    \{name\} \emph{is} in between \{name\} and \{name\},
    \emph{lives} on a floor between \{name\} and \{name\} \\
\hline
&
    \emph{Tourist sites}: Eiffel Tower, Big Ben, Colosseum, Statue of Liberty, Great Wall of China, Berlin Wall, Taj Mahal, Great Sphinx, Acropolis, Sagrada Familia, Machu Picchu, Grand Canyon &
    \emph{visited} before/after $\,^{(*)}$ &
    \emph{visited} \{site\} in between his \emph{visits} to\{site\} and \{site\}\\
 &
    \emph{Courses}: computer science, physics, algebra, analysis, economics, history, geography, robotics, philosophy, literature, chemistry, psychology &
    \emph{takes} before/after& 
    \emph{takes} \{course\} in between \{course\} and \{course\}\\
\multirow{5}{*}{\rotatebox{90}{\small Temporal}} & 
    \emph{Olympic disciplines:} fencing, climbing, archery, sprint, horseback riding, sailing, gymnastics, volleyball, karate, handball, swimming, rowing 
    & \emph{saw} before/after$\,^{(*)}$  
    & \emph{saw} \{discipline\} in between \{discipline\} and \{discipline\} \\
  &
    \emph{Relatives}: her mother, her father, her grandfather, her sister, her grandmother, her brother, her cousin, her husband, her uncle, her aunt, her nephew, her niece &
    \emph{talked to} before / after$\,^{(*)}$& 
    \emph{talked to} \{relative\} in between \emph{talking to} \{relative\} and \{relative\}\\
 &
    \emph{Stores}: the grocery store, the mall, the electronics store, the stationary store, the furniture store, the bookshop, the farmers' market, the pet store, the sports store, the department store,the garden center, the bakery &
    \emph{went to} before / after & 
    \emph{went to} \{store\} between \{store\} and \{store\}\\
 &
    \emph{Horse names}: Blue Monkey, Silver Saddle, Galloping Ghost, Quicksilver Quest, Velvet Storm, Majestic Moon, Starry Oasis, Radiant Runner, Thunder Roar, Midnight Blaze, Ruby Racer, Golden Phoenix &
    \emph{arrived} before / after$\,^{(*)}$ & 
    \{name\} \emph{arrived} between \{name\} and \{name\} \\
\hline
{\rotatebox{90}{\small Scalar}} &
    \emph{People}: James, David, Robert, Michael, William, Frank, Alice, Claire, Linda, Patricia, Emily, Grace &
    \emph{is}\,\,taller/shorter$\,^{(*)}$ \emph{than}, richer/poorer \emph{than}, faster/slower$\,^{(*)}$ \emph{than}, older/younger$\,^{(*)}$ \emph{than}, happier/sadder$\,^{(*)}$ \emph{than}, stronger/weaker \hspace{0.54cm} \emph{than}&
    \{name\}'s \emph{height/ wealth/ speed/ happiness level/ strength} is in between \{name\}'s and \{name\}'s  \emph{height/ wealth/ speed/ happiness level/ strength}, \{name\}'s \emph{age} is in between that of \{name\} and \{name\}\\
\hline
\end{tabular}

\end{table*}

\subsection{\WS{} finetuning train set}

The \WS{} finetuning train set consists of four scalar and four temporal skins that also occur in the \WS{} test set (in \cref{tab:supp-scenes}, they are denoted with *).
To increase diversity, we also created five new scalar and five new temporal skins.
All skins are populated with a new set of entities, that differ from the test set entities.
We list both the entities and new relations in \cref{tab:finetuning_entities}.

\begin{table*}[!h]
\caption{\textbf{Entities and relationships in the \WS{} finetuning train set.} In addition to the following entities/relationships, the finetuning dataset also contains the eight relationships that are denoted with an asterisk in \cref{tab:supp-scenes}, but using only the entities from this table. We do not repeat them here.}\label{tab:finetuning_entities}. 

    \scriptsize
    \renewcommand{\arraystretch}{1.5}
\begin{tabular}{|m{0.05\textwidth}<{\centering}|
m{0.45\textwidth}|
m{0.17\textwidth}|
m{0.22\textwidth}<{\centering}|}
\hline
\textbf{Domain} & \multicolumn{1}{c|}{\textbf{Entities}} & \multicolumn{1}{c|}{\textbf{Binary relations}} & \textbf{Ternary Relations} \\
\hline

\multirow{9}{*}{\rotatebox{90}{\small Temporal}} &\emph{Objects:} book on gardening, yoga mat, travel mug, desk lamp, phone case,
                cactus, yellow T-shirt, set of spoons, wall clock, scented candle, wine bottle, bag of cereals &
    \emph{bought} before / after &
    \emph{bought} \{object\} between the time when he \emph{bought} {\{object\}} and \{object\}\\

&\emph{People:} Nephew Alex, Niece Sarah, Sister Lisa, Brother Ben, Cousin Jake, Cousin Emily,
                       Uncle Rob, Aunt Annie, Uncle Chris, Aunt Marissa, Cousin Leo,Cousin Zoe &
    \emph{was born} before / after &
    \{person\} \emph{was born} between the birth of \{person\} and \{person\} \\
&\emph{Trees:} the maple tree, the apple tree, the pine, the cedar, the plum tree,the magnolia, the cherry tree, the oak, the willow, the cypress, the beech,the pear tree&
    \emph{planted} before / after & 
    \emph{planted} \{tree\} between the plantation of \{tree\} and \{tree\} \\
&\emph{Rooms:} the main bedroom, the bathroom, the kitchen, the guest bedroom, the dining room,the living room, the laundry room, the balcony, the staircase, the porch, the garage,the hallway &
    \emph{will renovate} before / after &
    \emph{will renovate} \{room\} between the renovation of \{room\} and \{room\} \\

&\emph{Objects}: the torn piece of fabric, the heavy footprints, the cryptic message, the blood-stained knife, the unexplained scratch, the hidden door, the secret diary, the broken violin, the burnt picture, the used bullet, the smelly cigar, the broken glass &
    \emph{discovered} before / after & 
    \emph{discovered} \{object\} between the time when he \emph{discovered} \{object\} and the time when he \emph{discovered}\{object\}\\

\hline

\multirow{9}{*}{\rotatebox{90}{\small Scalar \hspace{1.8cm}}} &\emph{Furniture}: the living room sofa,the wardrobe,the dining table,the king-sized bed,the large dresser,
                the master bookcase,the China cabinet,the buffet, the antique clock,the executive desk,
               the fridge &
    \emph{is} heavier/lighter than & 
    \{object\}'s \emph{weight} is in between \{object\}'s \emph{weight} and \{object\}'s \emph{weight}\\
&\emph{Mountains}: Frostfang Peak, Dragon's Breath Ridge, Lunarlight Summit, Crimson Crest, Mystshade Mount,
                    Eclipselock Spire, Whisperwind Crag, Thunder Throne, Crystalquill Range, Sirensong Cliff, Solstice Steeple, Ebonfire Crater &
    \emph{is} higher/lower than & 
    \{name\}'s \emph{height} is in between that of \{name\} and \{name\}\\
 &\emph{Objects}: box from Oregon, box from Missouri, box from Colorado, box from Pennsylvania, box from Arizona,
                    box from Louisiana, box from Michigan, box from Florida, box from Ohio, box from Texas,
                     box from Washington, box from Indiana &
    \emph{is} bigger/smaller than & 
    \emph{size} of the \{name\} is in between the \emph{sizes} of the \{name\} and the \{name\}\\
&\emph{Rivers}: the Amazon, the Seine, the Rhine, the Nile, the Danube, the Volga, the Mekong, the Hudson, the Yangtze, the Colorado, the Zambezi, the Loire &
    \emph{was} warmer/colder than & 
    \{name\}'s \emph{temperature} was in between that of \{river\} and \{river\}\\
&\emph{People}: the Smiths, the Browns, the Johnsons, the Andersons, the Thompsons, the Martins, the Robinsons, the Wilsons, the Thomases, the Taylors, the Davises, the Harrises &
    \emph{are} louder/quieter than & 
    are \{name\}'s \emph{noise level} between \{name\}' and \{name\}' \emph{noise levels} \\
\hline
\end{tabular}
\end{table*}

\subsection{\WS{} OOD test sets}\label{subsec:ood_testsets}

Apart from the regular test set, \WS{} also contains three test sets that are out-of-distribution with respect to the \WS{} finetuning training set.
First, we create a \emph{ood-size} set, which contains problems of size six, intended to test whether what has been learned during finetuning (which include only problems of size 3,4 and 5) extends to problems with more entities.
Second, we create an \emph{ood-query} where descriptions of the ordering of entities is as before, but we query about a different relation.
Instead of questions considering binary relationships, this test set contains questions with unary operators (min or max).
For instance it may query the truthvalue of ``\emph{the pyramid is the leftmost object}''. 
This is intended to test whether the finetuned models learn to reason about linear order, or just respond to our specific sets of questions.
Lastly, we create an \emph{ood-problem} test set, which contains a completely different set of problems that do not correspond to an underlying total order. 
In this new set of problems, we use a geometric setting with two binary operators: parallel and perpendicular, which can only be solved in a two-dimensional representation. 
This is intended to test whether finetuning on linear order problems can boost reasoning beyond linear order. 
Representative examples for all these test sets can be found in \cref{tab:ood_examples}.

\begin{table*}[h!]
\caption{\textbf{Representative samples for the problems in the three \WS{} OOD test sets.} In addition to its regular test set, the \WS{} bechmark has also three test sets that are out-of-distribution with respect to the \WS{} finetuning dataset: a \emph{ood-size} set, an \emph{ood-query} set, an \emph{ood-problem} set.
In this table, we give representative examples for each of these three test sets.}\label{tab:ood_examples}

\renewcommand{\arraystretch}{1.5}
\small
\begin{tabular}{|m{0.08\textwidth}<{\centering}|m{0.86\textwidth}|}
\hline
\textit{OOD set} & \multicolumn{1}{c|}{\textit{Sample}} \\
\hline
    \spheading{\small ood-size}  &  There are \textbf{6} people waiting in a line: Grace is in between Emily and Alice, Grace is behind Alice, Alice is in between William and Grace, Emily is in between Grace and Patricia and Patricia is in between Claire and Emily.
Is the following sentence `Claire is in between William and Patricia' TRUE or FALSE ?
Only respond with one of these 2 options: `TRUE', `FALSE' without any explanation. \\
    \hline
\spheading{\small ood-query }
& There is a stand with shelves vertically arranged. There are 3 objects, each placed on a unique shelf: the gray briefcase is below the silver watch and the silver watch is below the black book.
Is the following sentence `the black book is \emph{beneath all other objects}' TRUE or FALSE ?
Only respond with one of these 2 options: `TRUE', `FALSE' without any explanation. \\
\hline
\spheading{\small ood-problem }
& The map of the imaginary city of Gattaca looks like a grid, where all streets are straight and cross at a right angle. We consider 3 streets: Crick Avenue is perpendicular to Science Road and Crick Avenue is parallel to Darwin Drive.
Is the following sentence `Science Road is perpendicular to Darwin Drive' TRUE or FALSE ?
Only respond with one of these 2 options: `TRUE', `FALSE' without any explanation.  \\
\hline
\end{tabular}
\end{table*}




\section{Chain-of-thought prompts}\label{app:COT_prompts}

In one of the experiments reported in the main paper \cref{sec:results}, we used chain-of-thought (COT) prompting.
The exact prompt we used for the inference experiments is:

\begin{quote}
\textit{Reason step by step; write down your general strategy for solving this question; execute your steps; verify your reasoning, and revise strategy if this does not work. After doing all this reasoning, state your final answer with the sentence: `The final response is:' followed by one of these \textcolor{gray}{2} options: \textcolor{gray}{`TRUE', `FALSE'}, without adding anything more.}
\end{quote}

\noindent For the consistency and completeness experiments, we use the same prompt, but change the gray parts in the prompt above to match the possible responses for these two problem types.

\section{Numerical results}\label{appendix:acc_breakdown}
\begin{table*}[!h]
\caption{\textbf{Accuracies and response bias for the \WS{} Benchmark.} The results are shown across the three problem types (Inference, Consistency and Completeness detection) in both their Trivial and Normal versions, averaged across problem sizes 3, 4 and 5, for three chat-LLMs. Chance levels for accuracies is 50\%; biases are between -1 and 1; both given with 95\% confidence intervals.}\label{tab:supp_exp1}
\centering
\begin{tabular}{lccccccccccc}
\toprule
\multirow{1}{*}{Model} &   \multicolumn{2}{c}{ \underline{\;\;\;\;\;\;Inference\;\;\;\;\;\;}} & \multicolumn{2}{c}{\underline{\;\;\;\;Consistency\;\;\;\;}} & \multicolumn{2}{c}{ \underline{\;\;\;Completeness\;\;\;}} \\
&    {Trivial} &  { Normal} & {Trivial} &  {Normal} & {Trivial} & {Normal} \\
\midrule
\multicolumn{8}{c}{\underline{\it Accuracy}}\\

GPT3.5 & 64.8 {\tiny $\pm 1.1$} & 55.1 {\tiny $\pm 0.6$} & 52.2 {\tiny $\pm 0.7$} & 49.9 {\tiny $\pm 0.7$} & 59.9 {\tiny $\pm 1.0$} & 51.8 {\tiny $\pm 1.0$} \\
GPT4 & 90.3 {\tiny $\pm 0.7$} & 75.6 {\tiny $\pm 0.6$} & 71.2 {\tiny $\pm 0.6$} & 65.0 {\tiny $\pm 0.6$} & 93.1 {\tiny $\pm 0.6$} & 58.5 {\tiny $\pm 0.7$} \\
Llama2-chat & 62.2 {\tiny $\pm 0.8$} & 59.1 {\tiny $\pm 0.5$} & 54.2 {\tiny $\pm 0.4$} & 49.4 {\tiny $\pm 0.3$} & 60.5 {\tiny $\pm 0.7$} & 51.5 {\tiny $\pm 0.5$} \\
\hline
\multicolumn{8}{c}{\underline{\it Response Bias}}\\
GPT3.5 & 0.08 {\tiny $\pm 0.02$} & -0.34 {\tiny $\pm 0.01$} & -0.15 {\tiny $\pm 0.01$} & -0.00 {\tiny $\pm 0.01$} & -0.02 {\tiny $\pm 0.02$} & 0.20 {\tiny $\pm 0.02$} \\
GPT4 & -0.14 {\tiny $\pm 0.01$} & -0.20 {\tiny $\pm 0.01$} & -0.12 {\tiny $\pm 0.01$} & 0.00 {\tiny $\pm 0.01$} & 0.09 {\tiny $\pm 0.01$} & 0.78 {\tiny $\pm 0.01$} \\
Llama2-chat & -0.43 {\tiny $\pm 0.02$} & -0.63 {\tiny $\pm 0.01$} & 0.52 {\tiny $\pm 0.01$} & 0.79 {\tiny $\pm 0.01$} & 0.52 {\tiny $\pm 0.02$} & 0.83 {\tiny $\pm 0.01$} \\
\bottomrule
\end{tabular}

\end{table*}

To allow readers to run comparison with their models, we provide in Table \ref{tab:supp_exp1} the numbers corresponding to Figure \ref{fig:exp1}, breaking down the \WS{} benchmark by problem type both for accuracy and response bias.  In Table \ref{tab:supp_exp2}, we provide a similar breakdown for the prompting experiment, distinguishing the default prompting of the previous experiment (basic), the chain-of-thought prompting (COT) and two in-context learning conditions (IID and OOD). 

In Table \ref{tab:supp_exp3}, we present a similar breakdown for the finetuning experiment on a Llama2 70B model plus an additional 7B model (unreported in the main paper). 
The 7B experiment shows that when the model is too small, not only the initial accuracy is very low (all problem types close to chance level), but the finetuning does not help for all problem.
Indeed, only the normal inference and normal and trivial completeness problems benefit from finetuning, when 1M examples are used. 
The other problems stay at chance level, and more training examples only increase response bias even when tested with the same examples as in the training set (memorisation set). 
For the bigger 70B model in contrast, all of the problem types show evidence of learning and generalisation, although the consistency problems remain at a very low performance and relatively high response bias. 

Finally, in Table \ref{tab:exp3_external}, we report the results of the 70B finetuned model on three external benchmarks (MMLU, GSM8K and MATH). Because there were no improvements, we did not include the full results table in the main text, but we report them here for completeness.

\begin{table}[!h]
\caption{\textbf{Finetuned models' performance on external benchmarks.}}\label{tab:exp3_external}
    \centering
\begin{tabular}{lccc}
\toprule
&   Llama2$_{70B}$-FT+0K & Llama2$_{70B}$-FT+100K & Llama2$_{70B}$-FT+{1M} \\
\midrule
MMLU 5-shot (acc)      &   0.61 {\tiny $\pm 0.01$} & 0.61 {\tiny $\pm 0.01$} & 0.61 {\tiny $\pm 0.01$} \\
GSM8K 8-shot (pass@1)  &   0.51 {\tiny $\pm 0.03$} & 0.50 {\tiny $\pm 0.03$} & 0.51 {\tiny $\pm 0.03$} \\
MATH 4-shot (pass@1)   &   0.09 {\tiny $\pm 0.08$} & 0.09 {\tiny $\pm 0.08$} & 0.09 {\tiny $\pm 0.08$} \\
\bottomrule
\end{tabular}
\end{table}

\begin{table*}[!h]
\caption{\textbf{Accuracies and response bias for the prompting experiment.} 
The results are shown across the three problem types (Inference, Consistency and Completeness detection) in both their Trivial and Normal versions, averaged across problem sizes 3, 4 and 5, for three chat-LLMs for chain-of-thought (COT) and in-context learning (ICL) with either 5 in-distribution (IID) or out-of-domain (OOD) examples. Chance levels for accuracies is 50\%; biases are between -1 and 1; both given with 95\% confidence intervals.}\label{tab:supp_exp2}
\begin{tabular}{ll cc cc cc }
\toprule
\multirow{2}{*}{Model} &  & \multicolumn{2}{c}{ \underline{\;\;\;\;\;\;Inference\;\;\;\;\;\;}} & \multicolumn{2}{c}{\underline{\;\;\;\;Consistency\;\;\;\;}} & \multicolumn{2}{c}{ \underline{\;\;\;Completeness\;\;\;}} \\
&   & {Trivial} &  { Normal} & {Trivial} &  {Normal} & {Trivial} & {Normal} \\
\midrule

\multicolumn{8}{c}{\underline{\it Accuracy}}\\
\multirow[c]{4}{*}{GPT3.5} & basic & 64.8 {\tiny $\pm 1.1$} & 55.1 {\tiny $\pm 0.6$} & 52.2 {\tiny $\pm 0.7$} & 49.9 {\tiny $\pm 0.7$} & 59.9 {\tiny $\pm 1.0$} & 51.8 {\tiny $\pm 1.0$} \\
 & COT & 71.6 {\tiny $\pm 3.6$} & 59.3 {\tiny $\pm 2.0$} & 55.9 {\tiny $\pm 1.9$} & 50.7 {\tiny $\pm 2.1$} & 68.6 {\tiny $\pm 3.0$} & 53.6 {\tiny $\pm 3.4$} \\
 & ICL-IID & 56.2 {\tiny $\pm 3.5$} & 54.2 {\tiny $\pm 2.0$} & 49.0 {\tiny $\pm 2.0$} & 48.4 {\tiny $\pm 2.0$} & 56.6 {\tiny $\pm 2.6$} & 50.7 {\tiny $\pm 2.3$} \\
 & ICL-OOD & 65.9 {\tiny $\pm 3.5$} & 54.6 {\tiny $\pm 2.1$} & 51.0 {\tiny $\pm 2.0$} & 51.0 {\tiny $\pm 2.0$} & 53.8 {\tiny $\pm 2.5$} & 51.5 {\tiny $\pm 2.3$} \\
\hdashline

\multirow[c]{4}{*}{GPT4} & basic & 90.3 {\tiny $\pm 0.7$} & 75.6 {\tiny $\pm 0.6$} & 71.2 {\tiny $\pm 0.6$} & 65.0 {\tiny $\pm 0.6$} & 93.1 {\tiny $\pm 0.6$} & 58.5 {\tiny $\pm 0.7$} \\
 & COT & 90.3 {\tiny $\pm 2.2$} & 81.6 {\tiny $\pm 1.7$} & 76.0 {\tiny $\pm 1.8$} & 70.3 {\tiny $\pm 2.0$} & 94.9 {\tiny $\pm 1.6$} & 67.6 {\tiny $\pm 2.8$} \\
 & ICL-IID & 93.1 {\tiny $\pm 1.9$} & 73.3 {\tiny $\pm 1.9$} & 70.4 {\tiny $\pm 1.8$} & 61.7 {\tiny $\pm 1.8$} & 95.0 {\tiny $\pm 1.4$} & 62.3 {\tiny $\pm 2.6$} \\
 & ICL-OOD & 92.3 {\tiny $\pm 2.0$} & 71.1 {\tiny $\pm 1.9$} & 71.9 {\tiny $\pm 1.8$} & 65.0 {\tiny $\pm 1.7$} & 94.6 {\tiny $\pm 1.4$} & 63.8 {\tiny $\pm 2.6$} \\
\hdashline
\multirow[c]{4}{*}{Llama2-chat} & basic & 62.2 {\tiny $\pm 0.8$} & 59.1 {\tiny $\pm 0.5$} & 54.2 {\tiny $\pm 0.4$} & 49.4 {\tiny $\pm 0.3$} & 60.5 {\tiny $\pm 0.7$} & 51.5 {\tiny $\pm 0.5$} \\
 & COT & 69.3 {\tiny $\pm 3.2$} & 58.3 {\tiny $\pm 2.0$} & 50.8 {\tiny $\pm 1.8$} & 49.7 {\tiny $\pm 1.7$} & 60.0 {\tiny $\pm 3.0$} & 49.7 {\tiny $\pm 2.3$} \\
 & ICL-IID & 58.6 {\tiny $\pm 2.7$} & 58.8 {\tiny $\pm 1.5$} & 50.4 {\tiny $\pm 1.2$} & 50.2 {\tiny $\pm 1.1$} & 57.1 {\tiny $\pm 2.0$} & 51.4 {\tiny $\pm 1.2$} \\
 & ICL-OOD & 64.2 {\tiny $\pm 2.5$} & 57.6 {\tiny $\pm 1.4$} & 51.9 {\tiny $\pm 1.2$} & 49.3 {\tiny $\pm 1.0$} & 54.2 {\tiny $\pm 1.5$} & 50.8 {\tiny $\pm 1.0$} \\

\hline

\multicolumn{8}{c}{\underline{\it Response Bias}}\\

\multirow{4}{*}{{GPT3.5}} & basic &               0.08 {\tiny $\pm 0.02$} & -0.34 {\tiny $\pm 0.01$} &        -0.15 {\tiny $\pm 0.01$} & -0.00 {\tiny $\pm 0.01$} &    -0.02 {\tiny $\pm 0.02$} &     0.20 {\tiny $\pm 0.02$} \\
                               &    COT &              -0.07 {\tiny $\pm 0.07$} & -0.39 {\tiny $\pm 0.04$} &        -0.39 {\tiny $\pm 0.04$} & -0.22 {\tiny $\pm 0.04$} &    -0.11 {\tiny $\pm 0.06$} &     0.16 {\tiny $\pm 0.07$} \\
                               & ICL-IID &               0.15 {\tiny $\pm 0.08$} &  0.06 {\tiny $\pm 0.05$} &        -0.25 {\tiny $\pm 0.05$} & -0.11 {\tiny $\pm 0.05$} &     0.60 {\tiny $\pm 0.06$} &     0.69 {\tiny $\pm 0.05$} \\
                               & ICL-OOD &               0.15 {\tiny $\pm 0.07$} & -0.02 {\tiny $\pm 0.05$} &        -0.37 {\tiny $\pm 0.04$} & -0.27 {\tiny $\pm 0.04$} &     0.65 {\tiny $\pm 0.06$} &     0.71 {\tiny $\pm 0.05$} \\
\hdashline
\multirow{4}{*}{{GPT4}} & basic &              -0.14 {\tiny $\pm 0.01$} & -0.20 {\tiny $\pm 0.01$} &        -0.12 {\tiny $\pm 0.01$} &  0.00 {\tiny $\pm 0.01$} &     0.09 {\tiny $\pm 0.01$} &     0.78 {\tiny $\pm 0.01$} \\
                               &    COT &              -0.19 {\tiny $\pm 0.04$} & -0.20 {\tiny $\pm 0.03$} &        -0.17 {\tiny $\pm 0.04$} & -0.20 {\tiny $\pm 0.04$} &     0.05 {\tiny $\pm 0.03$} &     0.56 {\tiny $\pm 0.06$} \\
                               & ICL-IID &              -0.07 {\tiny $\pm 0.04$} & -0.16 {\tiny $\pm 0.04$} &        -0.28 {\tiny $\pm 0.04$} & -0.27 {\tiny $\pm 0.05$} &    -0.01 {\tiny $\pm 0.03$} &     0.65 {\tiny $\pm 0.05$} \\
                               & ICL-OOD &              -0.08 {\tiny $\pm 0.04$} & -0.10 {\tiny $\pm 0.04$} &        -0.19 {\tiny $\pm 0.04$} & -0.30 {\tiny $\pm 0.05$} &    -0.03 {\tiny $\pm 0.03$} &     0.64 {\tiny $\pm 0.06$} \\
\hdashline
\multirow{4}{*}{{Llama2-chat}} & basic &              -0.43 {\tiny $\pm 0.02$} & -0.63 {\tiny $\pm 0.01$} &         0.52 {\tiny $\pm 0.01$} &  0.79 {\tiny $\pm 0.01$} &     0.52 {\tiny $\pm 0.02$} &     0.83 {\tiny $\pm 0.01$} \\
                               &    COT &               0.00 {\tiny $\pm 0.08$} & -0.11 {\tiny $\pm 0.05$} &         0.20 {\tiny $\pm 0.05$} &  0.31 {\tiny $\pm 0.05$} &     0.43 {\tiny $\pm 0.06$} &     0.70 {\tiny $\pm 0.05$} \\
                               & ICL-IID &               0.04 {\tiny $\pm 0.09$} &  0.09 {\tiny $\pm 0.05$} &         0.37 {\tiny $\pm 0.05$} &  0.51 {\tiny $\pm 0.05$} &     0.70 {\tiny $\pm 0.06$} &     0.86 {\tiny $\pm 0.05$} \\
                               & ICL-OOD &              -0.01 {\tiny $\pm 0.09$} & -0.20 {\tiny $\pm 0.05$} &         0.18 {\tiny $\pm 0.06$} &  0.41 {\tiny $\pm 0.05$} &     0.75 {\tiny $\pm 0.06$} &     0.87 {\tiny $\pm 0.05$} \\


\bottomrule

\end{tabular}

\end{table*}

\begin{table*}[!h]
\caption{{\bf Accuracy and response bias for the finetuning experiment}. 
The results are shown across the Trivial and Normal versions of the three \WS{} problem types (Inference, Consistency and Completeness), for Llama2$_{7B}$ and Llama2$_{70B}$ models finetuned with instruction following data, plus increasing number of samples from the \WS{} training set (10k, 100k, 1M samples, each larger set containing smaller ones). Results are shown for the Memorization test set (identical to the 10k training set), the \WS{} test set (WS), and a size generalization test set with length 6 problems (ood-size). Chance levels for accuracies is 50\%; biases are between -1 and 1; both given with 95\% confidence intervals.
}\label{tab:supp_exp3}
\setlength{\tabcolsep}{2pt}
\centering

\begin{tabular}{ll cc cc cc}
\toprule
 & \multicolumn{1}{r}{Problem} & \multicolumn{2}{c}{\underline{\;\;\;\;\;Inference\;\;\;\;\;}}   & \multicolumn{2}{c}{\underline{\;\;\;\;Consistency\;\;\;\;}} & \multicolumn{2}{c}{\underline{\;\;\;Completeness\;\;\;}}  \\
Model & Test set &{ Trivial} & {Normal} & {Trivial} & {Normal} &  {Trivial} &  {Normal} \\
\hline
\multicolumn{8}{c}{\textit{\underline{Accuracy}}}\\

\multirow[c]{3}{*}{Llama2$_{7B}$-FT+0} & Memorization & 52.3 {\tiny $\pm 1.8$} & 52.4 {\tiny $\pm 0.9$} & 49.9 {\tiny $\pm 0.5$} & 49.7 {\tiny $\pm 0.5$} & 49.8 {\tiny $\pm 0.9$} & 49.8 {\tiny $\pm 0.7$} \\
 & WS & 51.7 {\tiny $\pm 1.1$} & 52.9 {\tiny $\pm 0.8$} & 49.3 {\tiny $\pm 0.7$} & 49.5 {\tiny $\pm 0.7$} & 50.6 {\tiny $\pm 1.0$} & 50.1 {\tiny $\pm 0.5$} \\
 & ood-size & 51.9 {\tiny $\pm 1.6$} & 51.1 {\tiny $\pm 1.0$} & 50.2 {\tiny $\pm 0.7$} & 49.2 {\tiny $\pm 0.7$} & 49.2 {\tiny $\pm 1.1$} & 49.8 {\tiny $\pm 1.1$} \\
\hdashline
\multirow[c]{3}{*}{Llama2$_{7B}$-FT+10k} & Memorization & 53.2 {\tiny $\pm 1.9$} & 53.2 {\tiny $\pm 0.9$} & 50.3 {\tiny $\pm 0.6$} & 50.9 {\tiny $\pm 0.7$} & 67.1 {\tiny $\pm 2.2$} & 50.2 {\tiny $\pm 0.3$} \\
 & WS & 51.7 {\tiny $\pm 1.5$} & 51.7 {\tiny $\pm 0.7$} & 50.2 {\tiny $\pm 0.5$} & 50.4 {\tiny $\pm 0.4$} & 62.5 {\tiny $\pm 1.9$} & 50.8 {\tiny $\pm 0.6$} \\
 & ood-size & 51.4 {\tiny $\pm 1.8$} & 52.1 {\tiny $\pm 1.1$} & 50.1 {\tiny $\pm 0.5$} & 50.0 {\tiny $\pm 0.5$} & 67.4 {\tiny $\pm 2.5$} & 51.2 {\tiny $\pm 1.0$} \\
\hdashline
\multirow[c]{3}{*}{Llama2$_{7B}$-FT+100k} & Memorization & 55.3 {\tiny $\pm 1.9$} & 60.8 {\tiny $\pm 1.3$} & 50.9 {\tiny $\pm 0.5$} & 50.4 {\tiny $\pm 0.5$} & 98.6 {\tiny $\pm 0.6$} & 56.4 {\tiny $\pm 1.6$} \\
 & WS & 52.5 {\tiny $\pm 1.3$} & 58.0 {\tiny $\pm 1.0$} & 50.3 {\tiny $\pm 0.4$} & 50.2 {\tiny $\pm 0.4$} & 91.9 {\tiny $\pm 1.4$} & 56.2 {\tiny $\pm 1.4$} \\
 & ood-size & 51.0 {\tiny $\pm 1.2$} & 55.0 {\tiny $\pm 1.2$} & 50.3 {\tiny $\pm 0.6$} & 49.9 {\tiny $\pm 0.5$} & 91.7 {\tiny $\pm 1.7$} & 53.6 {\tiny $\pm 1.5$} \\
\hdashline
\multirow[c]{3}{*}{Llama2$_{7B}$-FT+1M} & Memorization & 50.8 {\tiny $\pm 0.6$} & 68.2 {\tiny $\pm 1.4$} & 50.1 {\tiny $\pm 0.1$} & 50.2 {\tiny $\pm 0.2$} & 99.8 {\tiny $\pm 0.2$} & 92.6 {\tiny $\pm 1.7$} \\
 & WS & 50.2 {\tiny $\pm 0.3$} & 65.2 {\tiny $\pm 1.2$} & 50.6 {\tiny $\pm 0.4$} & 50.1 {\tiny $\pm 0.4$} & 92.7 {\tiny $\pm 1.3$} & 84.3 {\tiny $\pm 1.9$} \\
 & ood-size & 50.0 {\tiny $\pm 0.0$} & 58.8 {\tiny $\pm 1.4$} & 50.1 {\tiny $\pm 0.3$} & 50.2 {\tiny $\pm 0.2$} & 92.8 {\tiny $\pm 1.6$} & 79.0 {\tiny $\pm 2.5$} \\
 \hline
\multirow[c]{3}{*}{Llama2$_{70B}$-FT+0} & Memorization & 69.4 {\tiny $\pm 2.4$} & 56.7 {\tiny $\pm 1.4$} & 50.0 {\tiny $\pm 0.0$} & 50.0 {\tiny $\pm 0.0$} & 59.9 {\tiny $\pm 2.3$} & 49.9 {\tiny $\pm 1.2$} \\
 & WS & 63.7 {\tiny $\pm 0.8$} & 56.0 {\tiny $\pm 0.5$} & 50.0 {\tiny $\pm 0.0$} & 50.0 {\tiny $\pm 0.0$} & 59.9 {\tiny $\pm 0.7$} & 50.0 {\tiny $\pm 0.4$} \\
 & ood-size & 57.2 {\tiny $\pm 2.1$} & 53.9 {\tiny $\pm 1.4$} & 50.0 {\tiny $\pm 0.0$} & 50.0 {\tiny $\pm 0.0$} & 61.8 {\tiny $\pm 2.4$} & 50.5 {\tiny $\pm 1.3$} \\
\hdashline
\multirow[c]{3}{*}{Llama2$_{70B}$-FT+100k} & Memorization & 68.1 {\tiny $\pm 2.3$} & 80.6 {\tiny $\pm 1.4$} & 62.8 {\tiny $\pm 1.4$} & 57.6 {\tiny $\pm 1.2$} & 100.0 {\tiny $\pm 0.0$} & 99.3 {\tiny $\pm 0.5$} \\
 & WS & 67.0 {\tiny $\pm 0.8$} & 77.1 {\tiny $\pm 0.5$} & 60.8 {\tiny $\pm 0.4$} & 55.0 {\tiny $\pm 0.3$} & 97.4 {\tiny $\pm 0.3$} & 94.7 {\tiny $\pm 0.5$} \\
 & ood-size & 57.9 {\tiny $\pm 2.0$} & 68.3 {\tiny $\pm 1.6$} & 54.4 {\tiny $\pm 1.0$} & 51.9 {\tiny $\pm 0.8$} & 95.4 {\tiny $\pm 1.1$} & 87.9 {\tiny $\pm 2.2$} \\
\hdashline
\multirow[c]{3}{*}{Llama2$_{70B}$-FT+1M} & Memorization & 81.7 {\tiny $\pm 2.5$} & 82.8 {\tiny $\pm 1.4$} & 53.7 {\tiny $\pm 0.9$} & 52.2 {\tiny $\pm 0.7$} & 100.0 {\tiny $\pm 0.0$} & 100.0 {\tiny $\pm 0.0$} \\
 & WS & 79.7 {\tiny $\pm 0.9$} & 80.6 {\tiny $\pm 0.5$} & 54.9 {\tiny $\pm 0.3$} & 52.8 {\tiny $\pm 0.3$} & 98.5 {\tiny $\pm 0.2$} & 97.8 {\tiny $\pm 0.3$} \\
 & ood-size & 71.4 {\tiny $\pm 2.8$} & 74.0 {\tiny $\pm 1.8$} & 52.2 {\tiny $\pm 0.7$} & 50.7 {\tiny $\pm 0.6$} & 97.4 {\tiny $\pm 0.9$} & 95.8 {\tiny $\pm 1.4$} \\

\hline
\multicolumn{8}{c}{\textit{\underline{Response Bias}}}\\

\multirow[c]{3}{*}{Llama2$_{7B}$-FT+0} & Memorization & 0.36 {\tiny $\pm 0.08$} & 0.24 {\tiny $\pm 0.05$} & -0.75 {\tiny $\pm 0.03$} & -0.76 {\tiny $\pm 0.03$} & 0.91 {\tiny $\pm 0.03$} & 0.96 {\tiny $\pm 0.02$} \\
 & WS & 0.47 {\tiny $\pm 0.07$} & 0.38 {\tiny $\pm 0.04$} & -0.63 {\tiny $\pm 0.03$} & -0.57 {\tiny $\pm 0.04$} & 0.87 {\tiny $\pm 0.03$} & 0.94 {\tiny $\pm 0.03$} \\
 & ood-size & 0.33 {\tiny $\pm 0.09$} & 0.27 {\tiny $\pm 0.05$} & -0.63 {\tiny $\pm 0.04$} & -0.63 {\tiny $\pm 0.04$} & 0.74 {\tiny $\pm 0.06$} & 0.87 {\tiny $\pm 0.04$} \\
 \hdashline
\multirow[c]{3}{*}{Llama2$_{7B}$-FT+10k} & Memorization & -0.12 {\tiny $\pm 0.09$} & -0.26 {\tiny $\pm 0.05$} & 0.70 {\tiny $\pm 0.04$} & 0.69 {\tiny $\pm 0.04$} & 0.64 {\tiny $\pm 0.05$} & 0.99 {\tiny $\pm 0.01$} \\
 & WS & 0.11 {\tiny $\pm 0.08$} & -0.11 {\tiny $\pm 0.05$} & 0.84 {\tiny $\pm 0.02$} & 0.84 {\tiny $\pm 0.02$} & 0.69 {\tiny $\pm 0.04$} & 0.97 {\tiny $\pm 0.01$} \\
 & ood-size & -0.18 {\tiny $\pm 0.10$} & -0.47 {\tiny $\pm 0.05$} & 0.90 {\tiny $\pm 0.02$} & 0.89 {\tiny $\pm 0.03$} & 0.54 {\tiny $\pm 0.06$} & 0.95 {\tiny $\pm 0.02$} \\
 \hdashline
\multirow[c]{3}{*}{Llama2$_{7B}$-FT+100k} & Memorization & -0.62 {\tiny $\pm 0.06$} & -0.23 {\tiny $\pm 0.05$} & -0.40 {\tiny $\pm 0.05$} & -0.42 {\tiny $\pm 0.05$} & -0.01 {\tiny $\pm 0.01$} & 0.85 {\tiny $\pm 0.04$} \\
 & WS & -0.76 {\tiny $\pm 0.05$} & -0.39 {\tiny $\pm 0.04$} & -0.57 {\tiny $\pm 0.04$} & -0.58 {\tiny $\pm 0.04$} & -0.05 {\tiny $\pm 0.03$} & 0.84 {\tiny $\pm 0.03$} \\
 & ood-size & -0.79 {\tiny $\pm 0.06$} & -0.49 {\tiny $\pm 0.05$} & -0.59 {\tiny $\pm 0.05$} & -0.59 {\tiny $\pm 0.05$} & 0.03 {\tiny $\pm 0.04$} & 0.88 {\tiny $\pm 0.03$} \\
 \hdashline
\multirow[c]{3}{*}{Llama2$_{7B}$-FT+1M} & Memorization & -0.98 {\tiny $\pm 0.01$} & -0.41 {\tiny $\pm 0.04$} & 1.00 {\tiny $\pm 0.00$} & 1.00 {\tiny $\pm 0.00$} & -0.00 {\tiny $\pm 0.00$} & 0.13 {\tiny $\pm 0.03$} \\
 & WS & -1.00 {\tiny $\pm 0.01$} & -0.50 {\tiny $\pm 0.03$} & 0.90 {\tiny $\pm 0.02$} & 0.90 {\tiny $\pm 0.02$} & -0.15 {\tiny $\pm 0.03$} & 0.14 {\tiny $\pm 0.04$} \\
 & ood-size & -1.00 {\tiny $\pm 0.00$} & -0.55 {\tiny $\pm 0.04$} & 0.99 {\tiny $\pm 0.01$} & 0.99 {\tiny $\pm 0.01$} & -0.14 {\tiny $\pm 0.03$} & 0.20 {\tiny $\pm 0.06$} \\
 \hline
\multirow[c]{3}{*}{Llama2$_{70B}$-FT+0} & Memorization & 0.33 {\tiny $\pm 0.07$} & 0.30 {\tiny $\pm 0.04$} & 1.00 {\tiny $\pm 0.00$} & 1.00 {\tiny $\pm 0.00$} & 0.57 {\tiny $\pm 0.06$} & 0.81 {\tiny $\pm 0.05$} \\
 & WS & 0.57 {\tiny $\pm 0.02$} & 0.41 {\tiny $\pm 0.02$} & 1.00 {\tiny $\pm 0.00$} & 1.00 {\tiny $\pm 0.00$} & 0.63 {\tiny $\pm 0.02$} & 0.86 {\tiny $\pm 0.01$} \\
 & ood-size & 0.63 {\tiny $\pm 0.07$} & 0.29 {\tiny $\pm 0.05$} & 1.00 {\tiny $\pm 0.00$} & 1.00 {\tiny $\pm 0.00$} & 0.36 {\tiny $\pm 0.08$} & 0.69 {\tiny $\pm 0.07$} \\
 \hdashline
\multirow[c]{3}{*}{Llama2$_{70B}$-FT+100k} & Memorization & -0.63 {\tiny $\pm 0.05$} & -0.22 {\tiny $\pm 0.03$} & 0.60 {\tiny $\pm 0.03$} & 0.75 {\tiny $\pm 0.03$} & 0.00 {\tiny $\pm 0.00$} & 0.00 {\tiny $\pm 0.01$} \\
 & WS & -0.63 {\tiny $\pm 0.02$} & -0.29 {\tiny $\pm 0.01$} & 0.70 {\tiny $\pm 0.01$} & 0.83 {\tiny $\pm 0.01$} & -0.04 {\tiny $\pm 0.01$} & 0.05 {\tiny $\pm 0.01$} \\
 & ood-size & -0.80 {\tiny $\pm 0.04$} & -0.32 {\tiny $\pm 0.04$} & 0.81 {\tiny $\pm 0.03$} & 0.90 {\tiny $\pm 0.02$} & -0.08 {\tiny $\pm 0.02$} & 0.12 {\tiny $\pm 0.04$} \\
 \hdashline
\multirow[c]{3}{*}{Llama2$_{70B}$-FT+1M} & Memorization & 0.07 {\tiny $\pm 0.05$} & 0.05 {\tiny $\pm 0.03$} & 0.82 {\tiny $\pm 0.03$} & 0.88 {\tiny $\pm 0.02$} & 0.00 {\tiny $\pm 0.00$} & 0.00 {\tiny $\pm 0.00$} \\
 & WS & 0.26 {\tiny $\pm 0.02$} & 0.04 {\tiny $\pm 0.01$} & 0.84 {\tiny $\pm 0.01$} & 0.90 {\tiny $\pm 0.01$} & -0.02 {\tiny $\pm 0.00$} & 0.02 {\tiny $\pm 0.01$} \\
 & ood-size & 0.34 {\tiny $\pm 0.07$} & 0.12 {\tiny $\pm 0.04$} & 0.89 {\tiny $\pm 0.02$} & 0.93 {\tiny $\pm 0.02$} & -0.03 {\tiny $\pm 0.02$} & 0.04 {\tiny $\pm 0.03$} \\

 \bottomrule
\end{tabular}
\end{table*}

\begin{figure}[h!]
\includegraphics[width=0.57\textwidth]{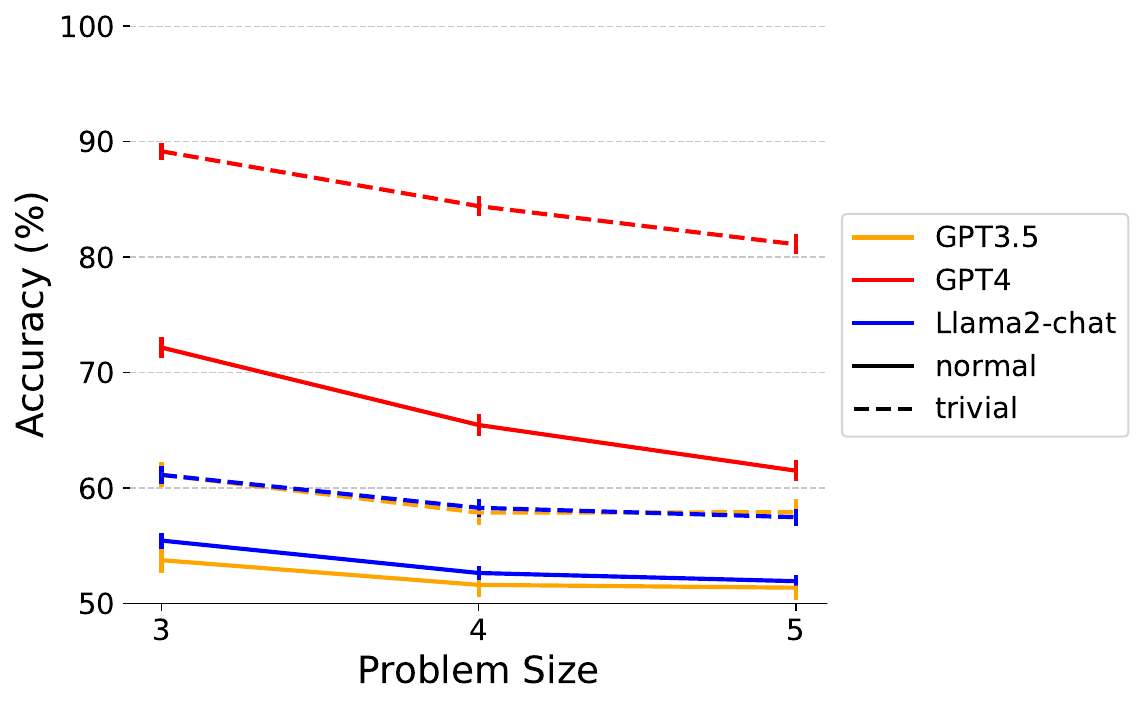}
\includegraphics[width=0.42\textwidth]{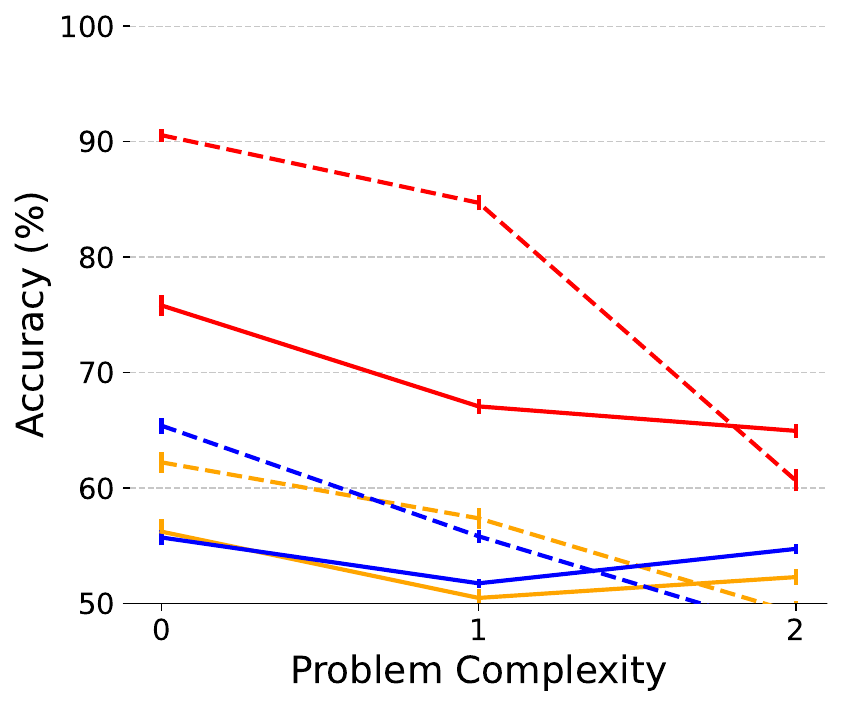}
\caption{\textbf{Effect of problem size (Left) and complexity (Right) on accuracy.} Results are averaged across the 3 \WS{} problem types (inference, consistency and completeness), separately for the normal and trivial versions. Problem size corresponds to the number of entities in the description, and complexity to the number of ternary relations involved in the problem (2 corresponding to 2 or more; note that due to logical constraints in problem generation this category only contains consistency problems and the normal version of inference problems). Chance levels for accuracies are 50\%, error bars represent 95\% confidence intervals.}
\label{fig:exp1_by_sizeco}
\end{figure}

\section{Detailed analyses}

In this section, we provide more detailed analysis of the \WS{} benchmark results, digging into the effects of problem size and problem complexity. The \WS{} test set contained problems varying in number of entities (problem size 3, 4, and 5).
\cref{fig:exp1_by_sizeco} (Left) shows that for both normal and trivial versions of the problems, models have a better performance for smaller problem sizes. 
For GPT4, it appears that the effect of problem size is bigger for the normal than the trivial version of the problem. \cref{fig:exp1_by_sizeco} (Right) shows that the model performance is also dependant on complexity, measured by the number of ternary relations mentioned in the entire problem (description and query). 
In \cref{tab:sup_monster_acc}, we provide the numerical values of the accuracies broken down both by problem size and complexity. Here, we separate the number of ternary relations in the description and in the query. 
Finally, we provide a breakdown of the accuracies by domain (scalar, spatial and temporal) in Table~\ref{tab:exp1_by_domain}. It can be seen that generally speaking, the scalar domain is the easiest and the temporal domain the hardest for GPT4 and Llama2-chat. 

\begin{table*}[!h]
\caption{\textbf{\WS{} accuracies split out by problem type, size and complexity.} Results for the 3 \WS{} problem types (Inference, Consistency and Completeness) and their trivial controls as a function of  description complexity (Cx.: 0, 1 or 2) and query arity (Qu.: binary vs. ternary), for different number of entities (3 to 5), across three chat-LLMs. Accuracies are given with 95\% confidence intervals, chance level at 50\%. 
}\label{tab:sup_monster_acc}

\setlength{\tabcolsep}{2pt}
\begin{tabular}{p{1.5cm}ll ccc  ccc  ccc}
\toprule
 &  &  & \multicolumn{3}{c}{\underline{\;\;\;\;\;GPT3.5\;\;\;\;\;}} & \multicolumn{3}{c}{\underline{\;\;\;\;\;\;GPT4\;\;\;\;\;\;}} & \multicolumn{3}{c}{\underline{\;\;\;Llama2-chat\;\;\;}} \\
Probl. & Cx. & Qu. & 3 & 4 & 5                   & 3 & 4 & 5                 &  3 & 4 & 5 \\
\hline
\multirow[c]{2}{1.5cm}{Infer. {\scriptsize (trivial)}} & \multirow[c]{2}{*}{c$_0$} & bin & 77.6 {\tiny $\pm 2.5$} & 71.0 {\tiny $\pm 2.7$} & 68.3 {\tiny $\pm 2.7$} & 98.6 {\tiny $\pm 0.7$} & 99.6 {\tiny $\pm 0.4$} & 98.8 {\tiny $\pm 0.6$} & 74.4 {\tiny $\pm 2.1$} & 63.8 {\tiny $\pm 1.9$} & 63.1 {\tiny $\pm 1.8$} \\
 &  & tern & 59.2 {\tiny $\pm 2.8$} & 56.2 {\tiny $\pm 2.7$} & 56.4 {\tiny $\pm 2.8$} & 92.8 {\tiny $\pm 1.5$} & 78.5 {\tiny $\pm 2.1$} & 73.5 {\tiny $\pm 2.1$} & 57.7 {\tiny $\pm 1.9$} & 58.2 {\tiny $\pm 1.8$} & 55.9 {\tiny $\pm 1.7$} \\
 \hdashline
\multirow[c]{6}{1.5cm}{Infer. {\scriptsize (normal)}} & \multirow[c]{2}{*}{c$_0$} & bin & 64.3 {\tiny $\pm 2.5$} & 58.3 {\tiny $\pm 2.5$} & 57.5 {\tiny $\pm 2.7$} & 96.6 {\tiny $\pm 1.0$} & 91.2 {\tiny $\pm 1.6$} & 83.5 {\tiny $\pm 2.1$} & 61.4 {\tiny $\pm 1.8$} & 56.7 {\tiny $\pm 1.5$} & 55.6 {\tiny $\pm 1.4$} \\
 &  & tern & 50.4 {\tiny $\pm 2.8$} & 49.8 {\tiny $\pm 2.8$} & 50.4 {\tiny $\pm 2.7$} & 90.8 {\tiny $\pm 1.7$} & 78.6 {\tiny $\pm 2.3$} & 69.7 {\tiny $\pm 2.4$} & 62.9 {\tiny $\pm 2.3$} & 56.4 {\tiny $\pm 2.1$} & 55.5 {\tiny $\pm 2.2$} \\
 & \multirow[c]{2}{*}{c$_1$} & bin & 54.0 {\tiny $\pm 2.6$} & 54.6 {\tiny $\pm 2.6$} & 54.2 {\tiny $\pm 2.6$} & 69.8 {\tiny $\pm 2.9$} & 79.2 {\tiny $\pm 2.5$} & 77.3 {\tiny $\pm 2.7$} & 58.9 {\tiny $\pm 1.7$} & 55.4 {\tiny $\pm 1.6$} & 54.5 {\tiny $\pm 1.5$} \\
 &  & tern & 64.3 {\tiny $\pm 2.7$} & 52.4 {\tiny $\pm 2.3$} & 51.3 {\tiny $\pm 2.6$} & 80.8 {\tiny $\pm 2.2$} & 67.9 {\tiny $\pm 2.3$} & 60.6 {\tiny $\pm 2.0$} & 76.0 {\tiny $\pm 2.2$} & 56.6 {\tiny $\pm 1.7$} & 52.8 {\tiny $\pm 1.4$} \\
 & \multirow[c]{2}{*}{c$_2$} & bin & 55.2 {\tiny $\pm 2.7$} & 50.3 {\tiny $\pm 2.6$} & 51.1 {\tiny $\pm 2.8$} & 71.5 {\tiny $\pm 2.9$} & 68.6 {\tiny $\pm 3.0$} & 58.4 {\tiny $\pm 3.1$} & 60.0 {\tiny $\pm 1.8$} & 55.9 {\tiny $\pm 1.7$} & 51.2 {\tiny $\pm 1.6$} \\
 &  & tern & 65.2 {\tiny $\pm 2.7$} & 52.9 {\tiny $\pm 2.8$} & 54.5 {\tiny $\pm 2.8$} & 82.1 {\tiny $\pm 2.2$} & 70.4 {\tiny $\pm 2.4$} & 64.3 {\tiny $\pm 2.1$} & 78.0 {\tiny $\pm 2.1$} & 62.0 {\tiny $\pm 2.2$} & 55.3 {\tiny $\pm 1.9$} \\
\hline

\multirow[c]{4}{1.5cm}{Consist. {\scriptsize (trivial)}} & c$_0$ & bin & 57.0 {\tiny $\pm 1.9$} & 52.7 {\tiny $\pm 2.0$} & 53.9 {\tiny $\pm 2.0$} & 82.7 {\tiny $\pm 1.4$} & 77.4 {\tiny $\pm 1.5$} & 72.1 {\tiny $\pm 1.5$} & 65.2 {\tiny $\pm 1.7$} & 61.4 {\tiny $\pm 1.5$} & 57.3 {\tiny $\pm 1.3$} \\
 & \multirow[c]{2}{*}{c$_1$} & bin & 58.5 {\tiny $\pm 2.7$} & 54.1 {\tiny $\pm 2.3$} & 51.8 {\tiny $\pm 2.3$} & 84.4 {\tiny $\pm 1.9$} & 81.7 {\tiny $\pm 1.7$} & 78.3 {\tiny $\pm 1.7$} & 62.6 {\tiny $\pm 2.0$} & 55.2 {\tiny $\pm 1.4$} & 53.7 {\tiny $\pm 1.1$} \\
 &  & tern & 50.4 {\tiny $\pm 2.9$} & 49.5 {\tiny $\pm 4.0$} & 51.1 {\tiny $\pm 4.8$} & 69.3 {\tiny $\pm 2.6$} & 60.6 {\tiny $\pm 3.9$} & 51.8 {\tiny $\pm 4.3$} & 49.6 {\tiny $\pm 1.0$} & 45.6 {\tiny $\pm 2.6$} & 47.4 {\tiny $\pm 2.8$} \\
 & c$_2$ & tern & 50.0 {\tiny $\pm 2.0$} & 48.0 {\tiny $\pm 2.0$} & 48.8 {\tiny $\pm 1.9$} & 69.1 {\tiny $\pm 1.8$} & 58.2 {\tiny $\pm 1.7$} & 54.6 {\tiny $\pm 1.7$} & 48.1 {\tiny $\pm 0.9$} & 47.3 {\tiny $\pm 1.2$} & 47.0 {\tiny $\pm 1.1$} \\
 \hdashline
\multirow[c]{6}{1.5cm}{Consist. {\scriptsize (normal)}} & \multirow[c]{2}{*}{c$_0$} & bin & 56.6 {\tiny $\pm 2.9$} & 51.7 {\tiny $\pm 3.0$} & 49.4 {\tiny $\pm 2.9$} & 78.6 {\tiny $\pm 2.2$} & 75.2 {\tiny $\pm 2.4$} & 69.6 {\tiny $\pm 2.4$} & 54.6 {\tiny $\pm 1.6$} & 51.2 {\tiny $\pm 1.2$} & 50.6 {\tiny $\pm 1.1$} \\
 &  & tern & 49.0 {\tiny $\pm 2.7$} & 47.7 {\tiny $\pm 2.9$} & 48.5 {\tiny $\pm 3.0$} & 75.3 {\tiny $\pm 2.4$} & 62.2 {\tiny $\pm 2.6$} & 57.2 {\tiny $\pm 2.6$} & 48.1 {\tiny $\pm 1.2$} & 48.0 {\tiny $\pm 1.1$} & 49.3 {\tiny $\pm 1.0$} \\
 & \multirow[c]{2}{*}{c$_1$} & bin & 50.3 {\tiny $\pm 2.7$} & 51.1 {\tiny $\pm 2.7$} & 50.4 {\tiny $\pm 2.9$} & 73.2 {\tiny $\pm 2.3$} & 69.1 {\tiny $\pm 2.6$} & 66.5 {\tiny $\pm 2.6$} & 50.0 {\tiny $\pm 0.7$} & 49.7 {\tiny $\pm 0.9$} & 49.6 {\tiny $\pm 0.9$} \\
 &  & tern & 49.9 {\tiny $\pm 2.9$} & 48.1 {\tiny $\pm 3.0$} & 49.7 {\tiny $\pm 2.9$} & 67.7 {\tiny $\pm 2.6$} & 54.5 {\tiny $\pm 2.6$} & 53.4 {\tiny $\pm 2.6$} & 47.8 {\tiny $\pm 1.4$} & 48.6 {\tiny $\pm 1.4$} & 48.8 {\tiny $\pm 1.3$} \\
 & \multirow[c]{2}{*}{c$_2$} & bin & 50.0 {\tiny $\pm 2.7$} & 50.0 {\tiny $\pm 2.9$} & 52.0 {\tiny $\pm 2.8$} & 73.3 {\tiny $\pm 2.4$} & 58.2 {\tiny $\pm 2.5$} & 56.3 {\tiny $\pm 2.2$} & 50.3 {\tiny $\pm 0.7$} & 50.4 {\tiny $\pm 1.0$} & 50.1 {\tiny $\pm 0.7$} \\
 &  & tern & 49.4 {\tiny $\pm 2.9$} & 48.8 {\tiny $\pm 2.7$} & 46.2 {\tiny $\pm 2.9$} & 68.0 {\tiny $\pm 2.5$} & 57.9 {\tiny $\pm 2.4$} & 55.5 {\tiny $\pm 2.5$} & 48.9 {\tiny $\pm 1.2$} & 45.4 {\tiny $\pm 1.5$} & 48.0 {\tiny $\pm 1.5$} \\
\hline

\multirow[c]{2}{1.5cm}{Complet. {\scriptsize (trivial)}} & \multirow[c]{2}{*}{c$_0$} & bin & 60.4 {\tiny $\pm 2.3$} & 59.3 {\tiny $\pm 2.4$} & 59.8 {\tiny $\pm 2.3$} & 98.0 {\tiny $\pm 0.8$} & 95.5 {\tiny $\pm 1.2$} & 92.1 {\tiny $\pm 1.4$} & 69.3 {\tiny $\pm 2.0$} & 67.4 {\tiny $\pm 2.1$} & 66.6 {\tiny $\pm 2.0$} \\
 &  & tern & 61.7 {\tiny $\pm 2.4$} & 58.5 {\tiny $\pm 2.4$} & 59.9 {\tiny $\pm 2.5$} & 92.8 {\tiny $\pm 1.5$} & 92.1 {\tiny $\pm 1.4$} & 88.2 {\tiny $\pm 1.7$} & 51.6 {\tiny $\pm 0.8$} & 53.5 {\tiny $\pm 1.0$} & 54.1 {\tiny $\pm 1.2$} \\
 \hdashline
\multirow[c]{2}{1.5cm}{Complet. {\scriptsize (normal)}} & \multirow[c]{2}{*}{c$_0$} & bin & 57.1 {\tiny $\pm 2.9$} & 54.8 {\tiny $\pm 3.0$} & 56.3 {\tiny $\pm 3.3$} & 67.9 {\tiny $\pm 2.5$} & 61.2 {\tiny $\pm 2.1$} & 58.5 {\tiny $\pm 2.1$} & 58.6 {\tiny $\pm 2.1$} & 56.1 {\tiny $\pm 2.0$} & 56.7 {\tiny $\pm 2.5$} \\
 &  & tern & 48.8 {\tiny $\pm 2.1$} & 51.0 {\tiny $\pm 2.1$} & 49.4 {\tiny $\pm 2.0$} & 59.1 {\tiny $\pm 1.4$} & 55.8 {\tiny $\pm 1.3$} & 54.8 {\tiny $\pm 1.2$} & 46.4 {\tiny $\pm 0.7$} & 50.0 {\tiny $\pm 0.2$} & 50.3 {\tiny $\pm 0.5$} \\

\bottomrule
\end{tabular}
\end{table*}

\begin{table*}[!h]
\caption{\textbf{\WS{} accuracies split out by domain.} Results are shown for three chat-LLMs, across the three \WS{} problem types and their trivial controls, broken down by the three domains (scalar, spatial and temporal). Accuracies are given with 95\% confidence intervals, chance level at 50\%.}\label{tab:exp1_by_domain}

\setlength{\tabcolsep}{2.1pt}
\begin{tabular}{l ccc ccc ccc}
\toprule
\multicolumn{1}{r}{Model}& \multicolumn{3}{c}{\underline{\;\;\;\;\;\;\;\;\;\;\;\;GPT3.5\;\;\;\;\;\;\;\;\;\;\;\;}} & \multicolumn{3}{c}{ 
 \underline{\;\;\;\;\;\;\;\;\;\;\;\;\;GPT4\;\;\;\;\;\;\;\;\;\;\;\;\;}} & \multicolumn{3}{c}{ \underline{\;\;\;\;\;\;\;\;\;Llama2-chat\;\;\;\;\;\;\;\;\;}} \\
\multicolumn{1}{c}{Probl. \;Domain} & \textit{scalar} & \textit{spatial} & \textit{temporal} & \textit{scalar} & \textit{spatial} & \textit{temporal} & \textit{scalar} & \textit{spatial} & \textit{temporal} \\
\midrule
\textit{Infer}. {\scriptsize (trivial)} & 68.1 {\tiny $\pm 2.0$} & 61.0 {\tiny $\pm 1.9$} & 65.4 {\tiny $\pm 1.9$} & 97.3 {\tiny $\pm 0.7$} & 88.3 {\tiny $\pm 1.3$} & 85.4 {\tiny $\pm 1.3$} & 72.8 {\tiny $\pm 1.6$} & 60.3 {\tiny $\pm 1.4$} & 53.6 {\tiny $\pm 0.8$} \\
\textit{Infer}. {\scriptsize (normal)} & 57.7 {\tiny $\pm 1.2$} & 53.7 {\tiny $\pm 1.1$} & 53.8 {\tiny $\pm 1.1$} & 78.4 {\tiny $\pm 1.0$} & 78.4 {\tiny $\pm 1.0$} & 70.1 {\tiny $\pm 1.1$} & 65.6 {\tiny $\pm 1.0$} & 57.1 {\tiny $\pm 0.7$} & 54.7 {\tiny $\pm 0.6$} \\
\hdashline
\textit{Consis}. {\scriptsize (trivial)} & 54.7 {\tiny $\pm 1.2$} & 51.4 {\tiny $\pm 1.1$} & 50.4 {\tiny $\pm 1.2$} & 77.0 {\tiny $\pm 0.9$} & 72.0 {\tiny $\pm 0.9$} & 64.8 {\tiny $\pm 1.0$} & 56.1 {\tiny $\pm 0.6$} & 56.7 {\tiny $\pm 0.9$} & 49.8 {\tiny $\pm 0.7$} \\
\textit{Consis}. {\scriptsize (normal)} & 51.0 {\tiny $\pm 1.2$} & 48.9 {\tiny $\pm 1.2$} & 49.9 {\tiny $\pm 1.2$} & 70.2 {\tiny $\pm 1.0$} & 63.9 {\tiny $\pm 1.0$} & 61.0 {\tiny $\pm 1.1$} & 50.3 {\tiny $\pm 0.2$} & 49.0 {\tiny $\pm 0.6$} & 48.8 {\tiny $\pm 0.6$} \\
\hdashline
\textit{Compl}. {\scriptsize (trivial)} & 59.3 {\tiny $\pm 1.6$} & 60.1 {\tiny $\pm 1.7$} & 60.4 {\tiny $\pm 1.7$} & 97.4 {\tiny $\pm 0.5$} & 92.6 {\tiny $\pm 1.0$} & 89.4 {\tiny $\pm 1.2$} & 62.8 {\tiny $\pm 1.3$} & 57.9 {\tiny $\pm 1.2$} & 60.7 {\tiny $\pm 1.2$} \\
\textit{Compl}. {\scriptsize (normal)} & 51.3 {\tiny $\pm 1.7$} & 52.4 {\tiny $\pm 1.8$} & 51.7 {\tiny $\pm 1.7$} & 69.4 {\tiny $\pm 1.5$} & 51.7 {\tiny $\pm 0.6$} & 54.3 {\tiny $\pm 0.9$} & 52.6 {\tiny $\pm 0.9$} & 51.6 {\tiny $\pm 0.8$} & 50.3 {\tiny $\pm 0.8$} \\
\hline
\it Avg. & 57.0 {\tiny $\pm 0.8$} & 54.6 {\tiny $\pm 0.8$} & 55.3 {\tiny $\pm 0.8$} & 81.6 {\tiny $\pm 0.6$} & 74.5 {\tiny $\pm 0.7$} & 70.8 {\tiny $\pm 0.7$} & 60.0 {\tiny $\pm 0.5$} & 55.4 {\tiny $\pm 0.5$} & 53.0 {\tiny $\pm 0.4$} \\

\bottomrule
\end{tabular}
\end{table*}

\newpage
\section{Examples prompts}
 
For convenience of the reader, we provide a fully spelled-out example prompts for our consistency experiment (\cref{subsec:consistency_query}), as well as a fully spelled-out ICL example (\cref{subsec:ICL_example}).

\subsection{Simple consistency query}\label{subsec:consistency_query}

{\bf Prompt:}\\
{\sffamily
\scriptsize
During the Olympics, Tim attended 3 sessions, each showcasing a unique discipline: Tim saw sailing before fencing and sprint before sailing.\\
Is the following sentence `Tim saw fencing after sprint' TRUE or FALSE ?\\
Only respond with one of these 2 options: `TRUE', `FALSE' without any explanation.\\
}
  
\subsection{In-context training with five examples from the same skin}\label{subsec:ICL_example}

{\bf Prompt:}\\
{\sffamily 
\scriptsize
In what follows, I am going to present you with a series of situations, each accompanied by a question. I will provide a few situations with the correct answer, as examples. I will leave only the last situation unanswered, and your task is to complete the series by providing the correct last answer.\\
------------------------\\
** Description of a new situation **\\
Over the past years, Bob visited 3 touristic places: Bob visited the Great Wall of China after the Sagrada Familia and the Sagrada Familia after the Berlin Wall.\\
Is the following sentence `Bob visited the Berlin Wall in between his visits to the Sagrada Familia and the Great Wall of China' TRUE or FALSE ?\\
Only respond with one of these 2 options: `TRUE', `FALSE' without any explanation.\\
 \verb!>>>! The correct answer is: FALSE\\
------------------------\\
** Description of a new situation **\\
Over the past years, Bob visited 3 touristic places: Bob visited the Acropolis before the Great Wall of China and the Grand Canyon after the Great Wall of China.\\
Is the following sentence `Bob visited the Grand Canyon before the Acropolis' TRUE or FALSE ?\\
Only respond with one of these 2 options: `TRUE', `FALSE' without any explanation.\\
 \verb!>>>! The correct answer is: FALSE\\
------------------------\\
** Description of a new situation **\\
Over the past years, Bob visited 3 touristic places: Bob visited the Eiffel Tower before the Statue of Liberty and the Big Ben after the Statue of Liberty.\\
Is the following sentence `Bob visited the Eiffel Tower before the Big Ben' TRUE or FALSE ?\\
Only respond with one of these 2 options: `TRUE', `FALSE' without any explanation.\\
 \verb!>>>! The correct answer is: TRUE\\
------------------------\\
** Description of a new situation **\\
Over the past years, Bob visited 3 touristic places: Bob visited the Great Sphinx after the Great Wall of China and the Sagrada Familia after the Great Sphinx.\\
Is the following sentence `Bob visited the Great Wall of China in between his visits to the Sagrada Familia and the Great Sphinx' TRUE or FALSE ?\\
Only respond with one of these 2 options: `TRUE', `FALSE' without any explanation.\\
 \verb!>>>! The correct answer is: FALSE\\
------------------------\\
** Description of a new situation **\\
Over the past years, Bob visited 3 touristic places: Bob visited the Sagrada Familia before the Great Sphinx and the Big Ben before the Sagrada Familia.\\
Is the following sentence `Bob visited the Great Sphinx in between his visits to the Sagrada Familia and the Big Ben' TRUE or FALSE ?\\
Only respond with one of these 2 options: `TRUE', `FALSE' without any explanation.\\
 \verb!>>>! The correct answer is: FALSE\\
------------------------\\
** Description of a new situation **\\
Over the past years, Bob visited 3 touristic places: Bob visited the Berlin Wall before the Great Wall of China and the Great Sphinx before the Berlin Wall.\\
Is the following sentence `Bob visited the Great Wall of China before the Great Sphinx' TRUE or FALSE ?\\
Only respond with one of these 2 options: `TRUE', `FALSE' without any explanation.\\
 \verb!>>>! The correct answer is:

\end{document}